\documentclass[]{onecat}
\usepackage{caption} 
\usepackage{hyperref}
\usepackage{cleveref}
\usepackage{verbatim}
\usepackage{amssymb} 
\usepackage{graphicx}
\usepackage{floatrow}
\usepackage{subcaption}
\usepackage{listings}
\usepackage{subcaption}
\usepackage{algorithm}
\usepackage{graphicx}
\usepackage{floatrow}
\usepackage{makecell}
\usepackage{enumitem}
\usepackage{wrapfig}  

\usepackage{subcaption} %

\usepackage[toc,page,header]{appendix}

\usepackage{pifont}

\definecolor{best}{rgb}{1.0, 0.85, 0.6}      
\definecolor{second}{rgb}{0.7, 0.9, 1.0}     
\definecolor{sh_blue}{rgb}{0,0.60,0.93}
\definecolor{sh_gray2}{rgb}{1,0.89,0.75}
\definecolor{lyellow}{rgb}{1,0.63,0.098}
\definecolor{lred}{rgb}{0.906,0.42,0.32}
\definecolor{color3}{rgb}{0.95,0.95,0.95}
\definecolor{mygray}{gray}{.9}
\definecolor{title}{rgb}{0.75,0.51,0.96}
\definecolor{genhaze}{rgb}{0.60, 0.57, 0.79}
\definecolor{bluegreen}{rgb}{0.44, 0.64, 0.77}
\definecolor{gray_venue}{rgb}{0.53,0.52,0.52}
\definecolor{color5}{rgb}{1,0.96,0.88}

\newlength{\Oldarrayrulewidth}

\lstdefinelanguage{json}{
  basicstyle=\ttfamily\small,
  breaklines=true,
  morecomment=[l]{//},
  morestring=[b]",
  stringstyle=\color{red!70!black},
  commentstyle=\color{gray!70!white},
  literate=
    *{0}{{{\color{gray}{0}}}}1
     {1}{{{\color{gray}{1}}}}1
     {:}{{{\color{black}{:}}}}1
     {,}{{{\color{black}{,}}}}1
     {\{}{{{\color{blue}{\{}}}}1
     {\}}{{{\color{blue}{\}}}}}1
     {[}{{{\color{blue}{[}}}}1
     {]}{{{\color{blue}{]}}}}1
}

\tcbuselibrary{listingsutf8} 
\tcbset{
    examplebox/.style={
        colback=green!5!white, 
        colframe=green!75!black, 
        coltitle=green!50!black, 
        fonttitle=\bfseries, 
        boxrule=0.5mm, 
        sharp corners, 
        enhanced, 
        attach boxed title to top left={
            yshift=-2mm, 
            xshift=5mm 
        },
        boxed title style={
            colframe=green!75!black, 
            colback=white, 
            sharp corners 
        }
    }
}

\newtcolorbox[auto counter, number within=section]{example}[2][]{%
    examplebox,
    title=Prompt   ~\thetcbcounter~(#2), 
    #1 
}

\usepackage{minitoc}

\title{\raisebox{-0.2cm}{\includegraphics[width=1.05cm]{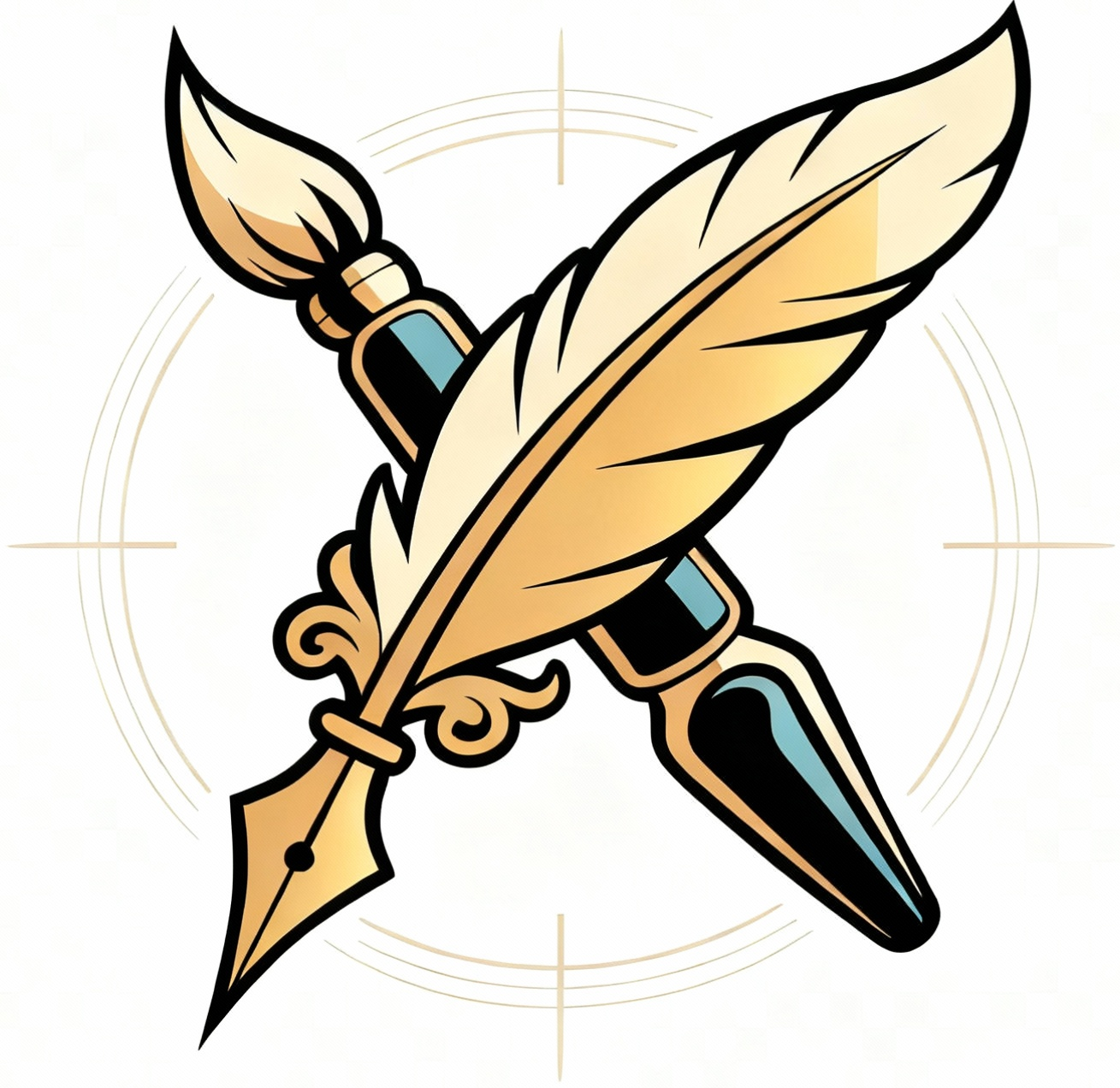}} \textit{\color{title}{PosterOmni}}: Generalized Artistic Poster Creation via Task Distillation and Unified Reward Feedback}

\author{%
\parbox{\textwidth}{\centering
Sixiang Chen$^{1,2*}$, Jianyu Lai$^{1,2*}$, Jialin Gao$^{2*}$, Hengyu Shi$^{2*}$, Zhongying Liu$^{2*}$,\\[2mm]
Tian Ye$^{1}$, Junfeng Luo$^{2}$, Xiaoming Wei$^{2}$, Lei Zhu$^{1,3\dagger}$
}}

\affiliation{%
\parbox{\textwidth}{\centering\small
$^1$The Hong Kong University of Science and Technology (Guangzhou), \quad
$^2$Meituan, \\ \quad
$^3$The Hong Kong University of Science and Technology
}}

\contribution[*]{Core Contribution}
\contribution[\dagger]{Corresponding Author}

\abstract{
Image-to-poster generation is a high-demand task requiring not only local adjustments but also high-level design understanding. Models must generate text, layout, style, and visual elements while preserving semantic fidelity and aesthetic coherence. The process spans two regimes: local editing, where ID-driven generation, rescaling, filling, and extending must preserve concrete visual entities; and global creation, where layout- and style-driven tasks rely on understanding abstract design concepts.  These intertwined demands make image-to-poster a multi-dimensional process coupling entity-preserving editing with concept-driven creation under image–prompt control.

To address these challenges, we propose PosterOmni, a generalized artistic poster creation framework that unlocks the potential of a base edit model for multi-task image-to-poster generation. PosterOmni integrates the two regimes, namely local editing and global creation, within a single system through an efficient data–distillation–reward pipeline, which includes:
(i) constructing multi-scenario image-to-poster datasets covering six task types across entity-based and concept-based creation;
(ii) distilling knowledge between local and global experts for supervised fine-tuning; and
(iii) applying unified PosterOmni Reward Feedback to jointly align visual entity-preserving and aesthetic preference across all tasks.
Additionally, we establish PosterOmni-Bench, a unified benchmark for evaluating both local editing and global creation. Extensive experiments show that PosterOmni significantly enhances reference adherence, global composition quality, and aesthetic harmony, outperforming all open-source baselines and even surpassing several proprietary systems.
}
\date{\today}
\checkdata[Project Page]{\url{https://ephemeral182.github.io/PosterOmni/}}

\begin{document}
\maketitle
\begin{figure}[hbt!]
    \centering
    \includegraphics[width=\textwidth]{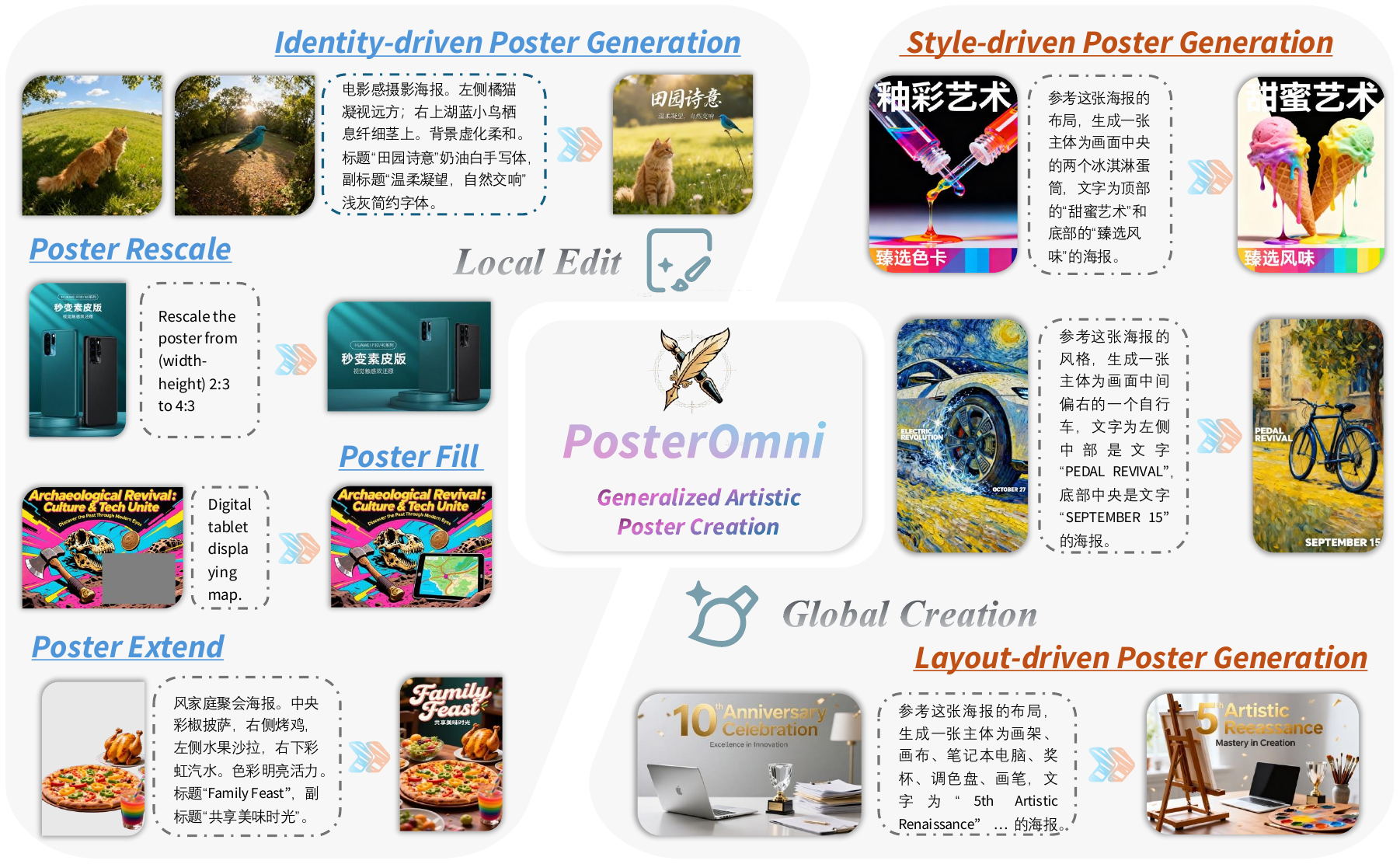}
    \caption{
        \textbf{PosterOmni unifies local editing and global creation within a single image-to-poster generation framework.}
It covers six representative tasks—extending, filling, rescaling, identity-driven, layout-driven, and style-driven poster generation—enabling the model to achieve both fine-grained visual editing and holistic aesthetic composition.}
    
    \label{fig:teaser}
\end{figure}
\section{Introduction}
Artistic poster generation is an important task in automated visual design. However, most real-world poster creation workflows remain image-centric—designers typically start from existing photographs, product images, or templates and transform them into complete visual posters. Such workflows require models capable of interpreting complex reference images, performing targeted modifications, and generating visually coherent results under both aesthetic and semantic constraints. Specifically, poster creators must not only modify image regions based on the given inputs but also adjust text, maintain layout–style balance, and preserve the design intent specified by the editing instruction.

Importantly, real-world poster creation involves two distinct forms. Designers may perform local adjustments that directly manipulate or preserve specific visual entities, or engage in global artistic creation that requires understanding abstract design concepts, such as layout or stylistic intent, to generate the scene holistically. These two regimes coexist in practical workflows, making image-to-poster creation a multi-dimensional problem that couples precise localized editing with concept-driven global transformation.

Nevertheless, no open framework currently targets multi-task image-to-poster creation. Existing open-source editing models, including Qwen-Image-Edit~\cite{qwenimage}, FLUX.1 Kontext~\cite{fluxkontext}, and ICEdit~\cite{icedit}, are strong natural-image editors (e.g., background replacement or object removal). While they can handle simple poster edits, they struggle with poster-specific creation. On tasks such as rescaling, identity-driven poster generation, or layout-driven global composition, these models frequently yield misaligned layouts, distorted text, or weakened aesthetic harmony. In contrast, commercial systems like Seedream-3/4~\cite{seedream3.0,seedream4.0}, GPT-Image~\cite{openai_gpt5_en}, and Gemini-2.5-Gen~\cite{gemini} handle such complex cases far better but are closed-source and costly to access at scale. This gap underscores the urgent need for an open image-to-poster framework that achieves accurate text rendering, reliable visual entity-preserving, and coherent layout/style understanding.

Our goal is to explicitly model the practical requirements of real-world poster creation. Therefore, different from previous mixed-training strategies of editing tasks, we revisit poster creation from a task-centric perspective and decompose image-to-poster generation into six representative tasks, which together span both reference-preserving local editing and concept-driven global creation:
\begin{itemize}
\item \textbf{Local Editing:} This family covers concrete modifications or generation guided by the input image, including Identity-driven Poster Generation, Poster Rescaling, Poster Filling, and Poster Extending. These tasks emphasize localized accuracy, spatial consistency, and faithful preservation of visual entities.
\item \textbf{Global Creation:} This family focuses on full-scene generation conditioned on higher-level design concepts. It includes Style-driven and Layout-driven Poster Generation, which require the model to reinterpret the poster holistically to achieve compositional harmony, stylistic coherence, and structural consistency.
\end{itemize}

Building on this formulation, we introduce \textbf{\textit{PosterOmni}}, a generalized artistic poster creation framework. Rather than being built from scratch, it leverages strong open-source editors and transforms them into specialized poster models through an efficient unified pipeline.
We first construct an automated data pipeline that generates high-quality, diverse data (PosterOmni-200K) covering six poster tasks for supporting training.
Following the decomposition of local editing and global creation, we innovatively perform task-distillation-based fine-tuning, integrating knowledge from expert models into a unified student network capable of precise local editing and holistic creation.
A dedicated unified PosterOmni Reward Model then provides general and task-specific signals to guide Diffusion-NFT to perform omni-edit reinforcement optimization, enabling targeted improvement across tasks.
Finally, we establish PosterOmni-Bench, a benchmark with paired (input, edit prompt) samples across multiple themes for consistent evaluation of local and global creation.
Experiments show that PosterOmni significantly improves image-to-poster generation performance, surpassing all open-source baselines and even several SOTA commercial systems.

Our main contributions are summarized as follows:
\begin{itemize}
    \item We design a fully automated data generation pipeline that produces high-quality, multi-scenario datasets across six poster tasks, ensuring balanced coverage of text and other visual elements variations.
    \item PosterOmni performs task distillation during the SFT stage, merging local and global experts into a unified lightweight student expert capable of learning both local editing and global generation.
    \item We propose a unified reward feedback stage by utilizing a unified PosterOmni Reward Model with the Omni-Edit RL stage, enabling general aesthetic and task-specific guidance that jointly optimizes local editing accuracy and global quality.
    \item We introduce the first comprehensive benchmark for multi-task image-to-poster generation, enabling consistent evaluation across diverse scenarios. PosterOmni achieves SOTA performance that surpasses all open-source models and rivals proprietary commercial systems.
\end{itemize}

\section{Related Works}
\noindent{\textbf{Image Editing.}}
Image editing aims to modify specific regions or attributes of an image while preserving other information~\cite{lin2025jarvisart,lin2025jarvisevo,icedit,qwenimage,bagel,gemini}.
Early diffusion-based methods relied on latent inversion~\cite{prompt2prompt,mokady2023null} or conditional guidance~\cite{controlnet,ipadapter}, but their flexibility and generalization remained limited.
Recent progress has shifted toward instruction-driven and multimodal editing frameworks, supported by more stable generative architectures such as flow-matching models~\cite{fluxkontext}, which improve controllability and fine-grained visual consistency.
Building on this foundation, ICEdit~\cite{icedit} and Step1X-Edit~\cite{step1x-edit} enhance localized, text-conditioned control, while Qwen-Image-Edit~\cite{qwenimage}, BAGEL~\cite{bagel}, and GPT-Image~\cite{openai_gpt5_en} integrate multimodal reasoning for more natural, instruction-following edits.
In this paper, distinct from these general editors, PosterOmni focuses on unified multi-task image-to-poster creation.

\noindent{\textbf{Artistic Poster Generation.}}
Poster generation~\cite{lin2023layoutprompter,yang2024posterllava,gao2025postermaker,chen2025posta,chen2025postercraft} is more challenging than generic image generation, as it requires coherent layout, typography, and visual storytelling.
Recent poster-focused studies have explored multiple paradigms.
{Text-to-poster} methods, such as POSTA~\cite{chen2025posta} and PosterCraft~\cite{chen2025postercraft}, typically treat poster design as a structured generation problem driven by textual intent, emphasizing design-aware composition, typography, and semantic alignment.
Complementary to this line, {layout-centric} approaches focus on generating or refining structured layouts as an intermediate representation (e.g., poster element arrangement, hierarchy, and alignment), including works along the direction of LayoutPrompter~\cite{lin2023layoutprompter} and PosterLayout~\cite{hsu2023posterlayout} that explicitly model layout planning to improve readability and visual balance.
Beyond poster-specific pipelines, recent diffusion-based models such as LayoutDiffusion~\cite{zheng2023layoutdiffusion}, TextDiffuser~\cite{textdiffuser}, and DesignDiffusion~\cite{wang2025designdiffusion} mainly focus on text-to-image generation, enhancing layout planning and text rendering but lacking flexible editing.
CreaiDesign~\cite{creatidesign}, PosterMaker~\cite{gao2025postermaker}, and DreamPoster~\cite{hu2025dreamposter} take initial steps toward image-to-poster generation by transforming normal images into poster-style outputs with added text, yet they do not address diverse poster tasks such as layout transfer, rescaling, or region filling.

Closed-source systems like GPT-Image~\cite{openai_gpt5_en}, Gemini 2.5-Flash~\cite{gemini}, and Seedream~\cite{seedream3.0,seedream4.0} demonstrate strong multimodal design capabilities for poster creation, but their training data, task coverage, and architecture remain opaque.
In contrast, PosterOmni targets the image-to-poster creation paradigm, unifying local editing and global composition through task distillation and unified reinforcement optimization, while covering a much broader range of poster editing and creation tasks than previous approaches.
Specifically, rather than addressing a single poster generation setting (pure text-to-poster or a fixed image-to-poster pipeline), PosterOmni expands the scope to a unified multi-task suite (e.g., layout transfer, rescaling with adaptive recomposition, and region filling) and provides an end-to-end workflow.

\section{Prerequisites for Flow Matching and Reinforce Learning} \label{sec:Prerequisites}

\noindent{\textbf{\textit{Flow Matching and Velocity Parameterization.}}}
Diffusion models~\cite{ho2020denoising,song2020score} generate samples by reversing a forward noising process, which can be written as a deterministic trajectory
\begin{equation}
x_t = \alpha_t x_0 + \sigma_t \epsilon,\quad \epsilon \sim \mathcal{N}(0,I),\; t \in [0,1],
\end{equation}
where $\alpha_t$ and $\sigma_t$ describe the evolution of the signal and noise, respectively. The velocity parameterization~\cite{zheng2023improved} predicts the tangent of this diffusion trajectory. Let

\begin{equation}
v = \dot{\alpha}_t x_0 + \dot{\sigma}_t \epsilon
\end{equation}
denote the instantaneous velocity along $x_t$. A neural network $v_\theta(x_t,t,c)$ is then trained to approximate this target field by minimizing
\begin{equation}
\mathbb{E}_{t,x_0,\epsilon}\bigl[w(t)\lVert v_\theta(x_t,t,c)-v\rVert_2^2\bigr],
\end{equation}
where $w(t)$ is a time-dependent weight. Sampling is performed by solving the deterministic ODE of the forward process:
\begin{equation}
dx_t = v_\theta(x_t,t,c)\, dt.
\end{equation}

Rectified flow~\cite{liu2022flow,lipman2022flow} can be viewed as a simplified instance of this velocity-parameterized formulation. Given a data sample $x_0 \sim X_0$ with condition $c$ and a Gaussian sample $x_1 \sim X_1$, it constructs the linear interpolation
\begin{equation}
x_t = (1-t)x_0 + t x_1,\quad t \in [0,1],
\end{equation}
whose velocity field satisfies $v = x_1 - x_0$. The corresponding flow-matching objective is
\begin{equation}
\mathcal{L}_{\text{FM}}(\theta)
= \mathbb{E}_{t,x_0,x_1}\left[\lVert v - v_\theta(x_t,t,c)\rVert_2^2\right].
\end{equation}
This setting is recovered from the diffusion trajectory by choosing $\alpha_t = 1-t$ and $\sigma_t = t$, which yields $v = \dot{\alpha}_t x_0 + \dot{\sigma}_t \epsilon = \epsilon - x_0$; identifying $x_1$ with $\epsilon$ recovers the rectified flow interpolation between $x_0$ and a Gaussian sample $x_1$.

\noindent{\textbf{\textit{Policy-Gradient Reinforce Learning for Diffusion Flows.}}}
Recent works~\cite{xue2025dancegrpo,liu2025flowgrpo,wang2025prefgrpo,li2025mixgrpo} formulate diffusion sampling as a multi-step Markov Decision Process (MDP), which enables the use of policy gradient methods such as PPO and GRPO. For rectified flows, however, the purely deterministic ODE dynamics prevent direct application of GRPO. FlowGRPO~\cite{liu2025flowgrpo} addresses this issue by introducing stochasticity through an SDE under the velocity parameterization:
\begin{equation}
dx_t = \Big[v_\theta(x_t,t) + \frac{g_t^2}{2t}\bigl(x_t + (1-t)v_\theta(x_t,t)\bigr)\Big]dt
       + g_t\, d\omega_t,
\end{equation}
where
\begin{equation}
g_t = a\sqrt{\frac{t}{1-t}}
\end{equation}
controls the magnitude of injected noise and $a$ is a tunable scale.

Discretizing this SDE with an Euler step of size $\Delta t$ yields a Gaussian transition kernel between adjacent states:
\begin{equation}
\begin{aligned}
\pi_\theta(x_{t-\Delta t}\mid x_t)
= \mathcal{N}\!\Bigl(
      x_t + \Big[v_\theta(x_t,t) +\\ \frac{g_t^2}{2t}\bigl(x_t + (1-t)v_\theta(x_t,t)\bigr)\Big]\Delta t,\;
      g_t^2 \Delta t\, I
    \Bigr).
\end{aligned}
\end{equation}
Such a parameterization makes the reverse-time transitions likelihood-tractable Gaussians, allowing existing policy gradient algorithms (e.g., GRPO) to be directly applied to diffusion models.

\noindent{\textbf{\textit{Diffusion Negative-aware Finetuning (DiffusionNFT).}}
DiffusionNFT~\cite{zheng2025diffusionnft} performs direct policy optimization on the forward diffusion process by leveraging a reward signal $r(x_0,c)\in[0,1]$. Rather than using standard policy gradient~\cite{liu2025flowgrpo,xue2025dancegrpo}, it forms a contrastive diffusion loss that pushes the model’s velocity predictor toward high-reward behavior and away from low-reward behavior.

Given an offline diffusion policy $v^{\text{old}}$, DiffusionNFT constructs implicit positive and negative policies:
\begin{equation}
v_\theta^{+}(x_t,t,c) = (1-\beta)v^{\text{old}}(x_t,t,c) + \beta v_\theta(x_t,t,c),
\end{equation}
\begin{equation}
v_\theta^{-}(x_t,t,c) = (1+\beta)v^{\text{old}}(x_t,t,c) - \beta v_\theta(x_t,t,c),
\end{equation}
where $\beta$ controls guidance strength. The training objective is
\begin{equation}
\mathcal{L}(\theta)=\mathbb{E}_{c,\pi^{\text{old}}(x_0|c),t}\left[
r\left\lVert v_\theta^{+}-v\right\rVert_2^2 + (1-r)\left\lVert v_\theta^{-}-v\right\rVert_2^2
\right],
\end{equation}
directly optimizing the new velocity field toward a reward-weighted improvement direction.
The reward is normalized as:
\begin{equation}
\begin{aligned}
r(x_0,c)
&= \tfrac12 +\\
&\tfrac12
   \operatorname{clip}\Bigg(
      \frac{
        r^{\text{raw}}(x_0,c)
        - \mathbb{E}_{\pi^{\text{old}}} r^{\text{raw}}(x_0,c) / Z_c
      }{1},
      -1,\, 1
   \Bigg).
\end{aligned}
\end{equation}

where $Z_c$ normalizes global reward scale.
Unlike policy-gradient diffusion RL, DiffusionNFT maintains forward consistency, integrates reinforcement signals implicitly into the velocity field, and entirely avoids likelihood approximation—enabling a simple, stable finetuning mechanism on the forward diffusion dynamics.
\begin{figure*}[!t]
    \centering
    \includegraphics[width=16.5cm]{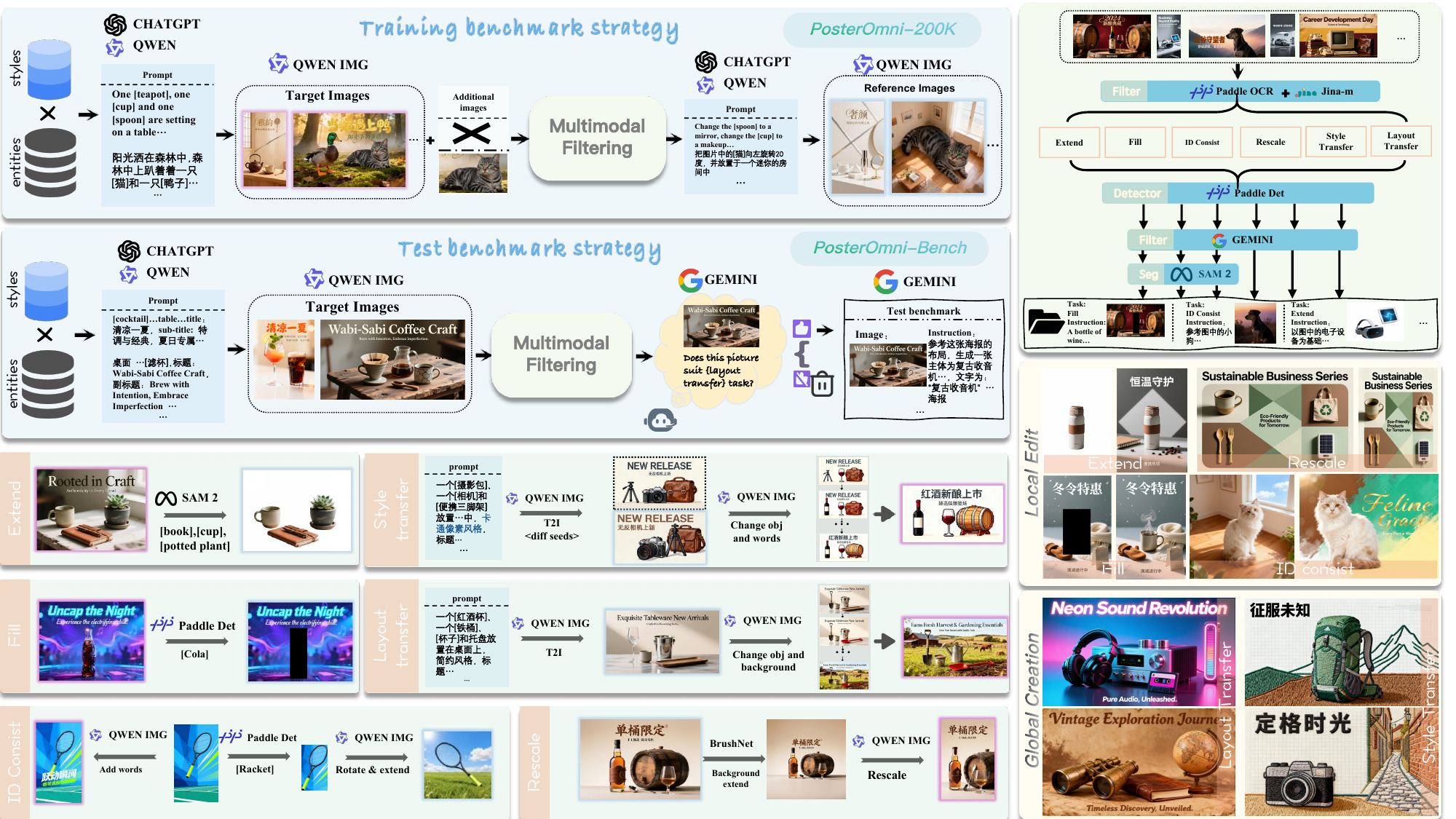}
    \caption{We decompose image-to-poster generation into local editing and global creation, including extending, filling, rescaling, identity-driven, layout-driven, and style-driven generation.
Our overall pipeline integrates prompt generation, image generation, multimodal filtering, and task-specific construction into a unified framework for large-scale, image-to-poster data generation.
We then propose \textbf{PosterOmni-200K} and \textbf{PosterOmni-Bench}, which encompass six major poster themes and multi-image input scenarios.}
    \label{fig:data pipeline}
\end{figure*}

\section{PosterOmni Pipeline}
\subsection{Automated Data Construction}\label{data construction}
To enable unified learning across diverse image-to-poster creation,
we develop an automated data construction pipeline that synthesizes large-scale, task-aligned paired datasets without manual annotation.
As illustrated in Fig.\ref{fig:data pipeline}, the pipeline integrates prompt generation, image generation, and multimodal filtering into a unified framework, constructing task-specific input–output pairs that ultimately form \textbf{PosterOmni-200K} and \textbf{PosterOmni-Bench}, supporting both fine-tuning and final evaluation.

\noindent{\textbf{\textit{Prompt and Image Generation:}}} To construct diverse, high-quality image-to-poster data, we first generate large-scale (prompt, image) pairs with rich typographic and stylistic variation. We sample combinations of entities (e.g., products, food, events) and styles (e.g., minimalist, vintage, Y2K) from curated libraries to form structured prompts. Using GPT~\cite{openai_gpt5_en} and Qwen3~\cite{yang2025qwen3}, we produce fluent descriptions that reflect real poster themes and specify layout, context, and aesthetic intent. Qwen-Image~\cite{qwenimage} and other SOTA generator~\cite{flux} then render multiple candidate images per prompt. Finally, early filtering removes samples with missing subjects, corrupted text, or collapsed layouts.

\noindent{\textbf{\textit{Multimodal Filtering:}}}
After generating initial text-to-image pairs, we apply multimodal filtering to ensure data quality and task alignment. For the PosterOmni-200K training set, each sample undergoes multi-stage verification with PaddleOCR~\cite{cui2025paddleocr} and Jina-clip-v2~\cite{jina-clip-v2} to check textual correctness and layout–content consistency. This removes samples with mismatched captions, misplaced typography, or low visual–textual coherence, ensuring semantic fidelity and aesthetic quality. For the PosterOmni-Bench, we adopt stricter filtering. In addition to OCR-based checks, Gemini-2.5-Flash~\cite{gemini} evaluates task suitability (e.g. whether an image contains an analyzable layout for layout-driven tasks). We further apply SAM-2~\cite{sam2} for segmentation-based refinement, generating localized regions or masks as supervision targets for task-specific editing.

\begin{wrapfigure}{r}{8.5cm}
    \centering
    \includegraphics[width=8.3cm]{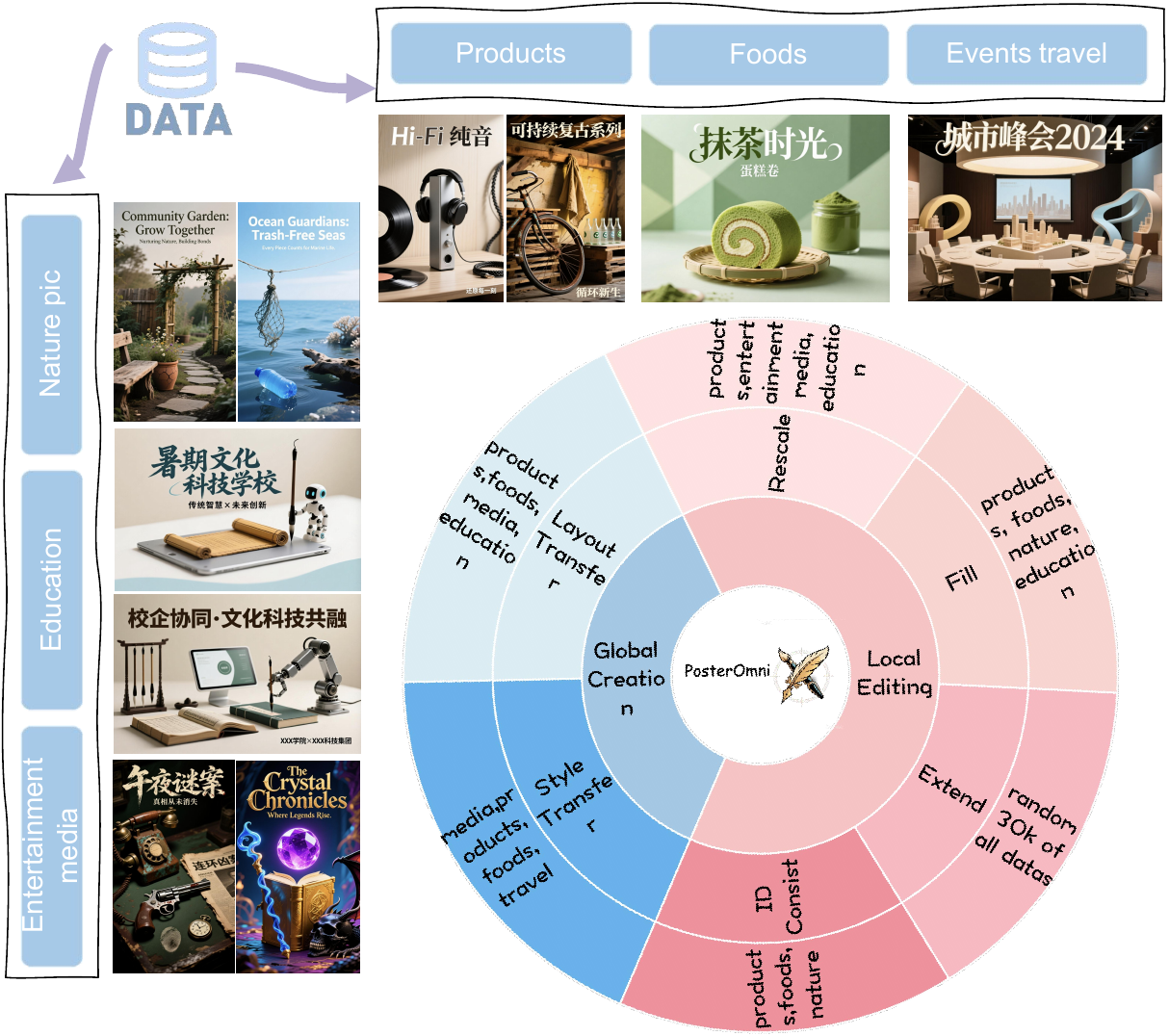}
    \caption{\textbf{PosterOmni datasets} cover six poster themes (products, foods, events/travel, nature, education, and entertainment) and support both local editing and global creation tasks.}
    \label{fig:pie}
\end{wrapfigure}

\noindent{\textbf{\textit{Task-Specific Image-to-Poster Construction:}}}
Building on the filtered text-to-image corpus, we construct paired image-to-poster samples covering six tasks—extending, filling, rescaling, ID-driven, layout-driven, and style-driven generation—capturing spatial completion, aspect-ratio adjustment, subject preservation, layout transformation, and aesthetic adaptation. Each task is implemented through a modular pipeline: extending/filling use SAM2-based masking, rescaling applies BrushNet~\cite{ju2024brushnet}, ID-driven uses PaddleDet~\cite{paddledet} and strong edit models, and layout/style-driven tasks rely on prompt-controlled re-rendering. The resulting PosterOmni-200K dataset contains over 200K paired samples with diverse supervision across these tasks. For evaluation, PosterOmni-Bench provides manually curated prompts and images. All datasets span six major poster themes—Products, Food, Events, Nature, Education, and Entertainment (Fig.~\ref{fig:pie})—supporting consistent assessment of both local editing and global creation. More construction details can be found in our supp..

\subsection{PosterOmni Training Workflow}
Given a foundation editor $M_{\text{base}}$, our objective is to train a model $M_{\text{omni}}$ that can support performing precise poster \textit{local editing} and \textit{global creation} across six representative tasks:
\begin{equation}
\mathcal{T} = 
\underbrace{\{\text{Rescaling, Filling, Extending, ID}\}}_{\text{Local Editing}},
\underbrace{\{\text{Style, Layout}\}}_{\text{Global Creation}}.
\tag{1}
\end{equation}

To achieve this goal, we design a framework that evolves from task-specific fine-tuning to task distillation and unified reward-guided reinforcement learning, as shown in Fig.\ref{fig:overview}. Task distillation unifies different image-to-poster generation abilities into a single backbone for diverse image-to-poster tasks, while the unified PosterOmni Reward Model provides general and task-specific signals to guide DiffusionNFT–based RL, improving the task-specific performance and aligning general human preference.

\begin{figure*}[!t]
    \centering
    \includegraphics[width=16.5cm]{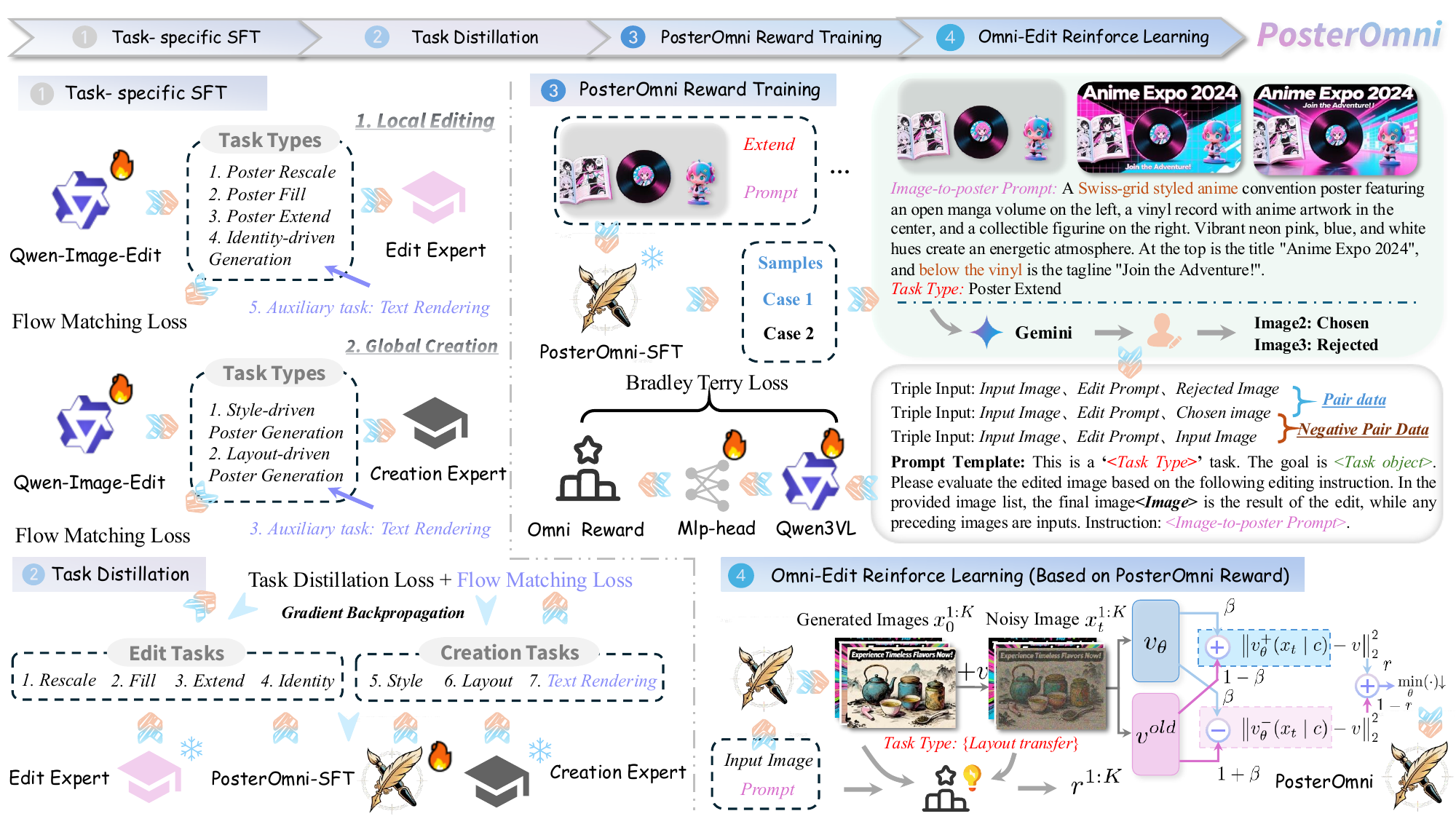}
    \caption{\textbf{PosterOmni training workflow} through four stages: (i) task-specific SFT for local and global experts, (ii) task distillation to integrate them into a single PosterOmni-SFT model, (iii) reward training for the unified PosterOmni Reward $R_{\text{omni}}$, and (iv) Omni-Edit RL using DiffusionNFT to align creation with human-preferred aesthetics and precision. For clarity, only one task is illustrated in (iii) and (iv).}
    \label{fig:overview}
\end{figure*}

\noindent{\textbf{\textit{Task-Specific Supervised Fine-Tuning.}}}
PosterOmni first performs task-specific fine-tuning on the base editor $M_{\text{base}}$ to establish a foundation for unified optimization. Instead of training all tasks jointly, we divide them into two different groups—local editing $\mathcal{T}_{\text{local}}=\{\text{Rescaling, Filling, Extending, Identity-driven}\}$ and global creation $\mathcal{T}_{\text{global}}=\{\text{Style-driven, Layout-driven}\}$. Local editing focuses on precision and reference entity consistency, while global creation emphasizes abstract layout and style understanding and generation. This decomposition reduces interference between pixel-level correction and high-level composition, yielding two specialized experts $E_{\text{local}}$ and $E_{\text{global}}$.
Each task $t\in\mathcal{T}$ is optimized using paired data ($I_{\text{in}}, p_t, I_{\text{out}}$) under the flow-matching loss:
\begin{equation}
\mathcal{L}_{\text{SFT}}=
\mathbb{E}_{x_t,v_t\sim q(x_t,v_t)}
\left[|v_t - v_\theta(x_t,t,c_t)|_2^2\right],
\end{equation}
where $v_\theta$ is predicted velocity field and $c_t=(I_{\text{in}},p_t)$ is conditioning input. To maintain efficiency and preserve base model’s ability, fine-tuning LoRA is applied in this stage.

Beyond the six tasks, an auxiliary text-rendering objective is introduced to preserve text generation. We build a text-only dataset—images containing only textual content without layout or style—and mix these samples into both local and global SFT phases. This maintains character-level rendering quality and prevents degradation during specialization, resulting in two robust experts.

\noindent{\textbf{\textit{Task Distillation.}}} 
The objective of PosterOmni is to build a unified model that can simultaneously handle image-to-poster tasks with both precision and global understanding. After obtaining two experts $E_{\text{local}}$ and $E_{\text{global}}$, the challenge lies in integrating their abilities without mutual interference. A straightforward approach is to merge LoRA adapters via linear addition, SVD-based fusion, or ZipLoRA compression~\cite{shah2024ziplora}. However, these parameter-level methods directly fuse both experts within a single latent space, and the disparity between local editing and global creation often causes severe degradation.

Inspired by knowledge distillation, we design a task distillation framework where a new student expert learns under the joint supervision of $E_{\text{local}}$ and $E_{\text{global}}$. Instead of merging parameters, the student progressively acquires their crucial knowledge, forming a unified backbone for diverse image-to-poster tasks. This approach offers key advantages over common mixed-task joint training:
(i) each expert specializes in its own domain and characteristic, avoiding destructive interference.
(ii) the student receives consistent teacher signals, accelerating convergence; and
(iii) the decoupled expert structure simplifies data organization without extensive task balancing.

Formally, the training objective combines an auxiliary text-rendering loss with the main task distillation loss. The auxiliary term preserves text-rendering ability for visual–textual consistency during the distillation process, while the main loss aligns the student with both expert guidance. Specifically, it is defined as:
\begin{equation}
\begin{aligned}
\mathcal{L}_{\text{total}} 
&=
\underbrace{
\mathbb{E}_{x_t,v_t\sim q(x_t,v_t)}
\left[\|v_t - v_\theta(x_t,t,c_t)\|_2^2\right]
}_{\text{Auxiliary (Text Rendering) Loss}} 
\\
&\quad + 
\underbrace{
\lambda_{\text{E}}\,
\mathbb{E}_{x_t,v_t\sim q(x_t,v_t)}
\left[\|v_\theta(x_t,t,c_t) - v_{\text{E}}(x_t,t,c_t)\|_2^2\right]
}_{\text{Task Distillation Loss}}
\end{aligned}
\end{equation}
where $v_\theta$ denotes the student’s predicted velocity field and $v_{\text{E}}$ denotes the expert output for the corresponding task. Through this process, PosterOmni integrates both task types into a unified backbone. The resulting $M_{\text{sft}}$ inherits the precision of local experts and the generative reasoning of global experts, forming a solid foundation for RL stage.

\noindent{\textbf{\textit{PosterOmni Reward Training.}}}
Image-to-poster requires balancing local entities' precision, abstract composition, and aesthetic preference. While supervised fine-tuning enables task performance, it often leads to shortcut learning and poor generalization, while limiting higher-level aesthetic understanding. To overcome this, we introduce the unified PosterOmni Reward Model $R_{\text{omni}}$, which provides both general and task-specific reward signals to align the model with human preferences in aesthetics and editing precision across diverse poster tasks.

To train $R_{\text{omni}}$, we build a preference dataset from outputs of the SFT-trained PosterOmni model. For each image-to-poster prompt, paired results are generated and filtered by Gemini-2.5-Pro~\cite{gemini}, after which annotators choose the more aesthetic and task-faithful one. We also add a novel negative-pair strategy, treating the input as the rejected sample and the output result as the preferred one to encourage meaningful image-to-poster judgement. Importantly, differences between pairs often stem from two complementary aspects: one is global aesthetic appeal (e.g., text rendering, color balance), while the others diverge in their adherence to the instruction or task type. This enables $R_{\text{omni}}$ to jointly learn both aesthetic and task-specific quality dimensions.
Each sample forms a quadruplet $(I_{\text{in}}, p_t, \text{edit}, I_{\text{chosen}}, I_{\text{rejected}})$.
Built on the Qwen3VL~\cite{yang2025qwen3} encoder with a lightweight MLP head, $R_{\text{omni}}$ jointly encodes visual quality and instruction with its task type for unified evaluation.
Preference alignment follows the Bradley–Terry formulation, converting pairwise comparisons into a differentiable objective:

\begin{equation}
\mathcal{L}_{\text{BT}} =
-\mathbb{E}_{(I_{\text{chosen}}, I_{\text{rejected}})}
\Big[
\log \sigma \big( r_\theta(I_{\text{chosen}}) - r_\theta(I_{\text{rejected}}) \big)
\Big],
\end{equation}
where $r_\theta(\cdot)$ denotes the predicted scalar reward and $\sigma(\cdot)$ ensures probabilistic ranking consistency. More data construction and training details can be found in supp..

\noindent{\textbf{\textit{Omni-Edit Reinforcement Learning.}}}
Recent advances like DiffusionNFT~\cite{zheng2025diffusionnft} reformulate reinforcement learning for diffusion models by optimizing policies along the forward process instead of the reverse trajectory used in GRPO~\cite{xue2025dancegrpo,liu2025flowgrpo}. This stabilizes gradients and allows continuous reward modulation in the forward direction. Building on this idea, we first extend DiffusionNFT to image-to-poster generation and integrate it with our unified reward model $R_{\text{omni}}$, forming the Omni-Edit RL strategy.

Unlike UniWorld-V2~\cite{uniworldv2}, which scales multimodal LLMs and uses logits as generic editing rewards, our method couples DiffusionNFT with task-specific scores from $R_{\text{omni}}$, enabling joint optimization of local and global poster creation while improving poster-specific aesthetic quality. Using unified rewards from $R_{\text{omni}}$, we further refine the diffusion model via a DiffusionNFT-based flow-matching update. Instead of conventional policy gradients, PosterOmni injects reward signals directly into the forward diffusion objective, guiding the model toward high-reward edits and away from low-reward ones. The policy loss is formulated as:
\begin{equation}
\mathcal{L}_{\text{RL}}
=\mathbb{E}_{c,t}\Big[
r,|v_\theta^{+}(x_t,c,t)-v|_2^2
+(1-r),|v_\theta^{-}(x_t,c,t)-v|_2^2
\Big],
\end{equation}
where $v$ denotes the target velocity field and $r\in[0,1]$ is the normalized reward derived from $R_{\text{omni}}$.
The positive and negative policies are defined as:
\begin{equation}
\begin{aligned}
v_\theta^{+}(x_t,c,t)=(1-\beta)v_{\text{old}}(x_t,c,t)+\beta v_\theta(x_t,c,t),&\\
v_\theta^{-}(x_t,c,t)=(1+\beta)v_{\text{old}}(x_t,c,t)-\beta v_\theta(x_t,c,t),
\end{aligned}
\end{equation}
where $\beta$ controls the update strength between current and previous policies. This contrastive objective aligns the model’s velocity field with human-preferred aesthetics while preserving diffusion consistency. Through the Omni-Edit RL stage, PosterOmni boosts precise local editing and global reasoning, while achieving human-aligned aesthetic optimization for visual quality. For further theoretical reasoning and explanation, please refer to the suppl..

\section{Experiment}
\begin{figure*}[!t]
    \centering
    \includegraphics[width=16.5cm]{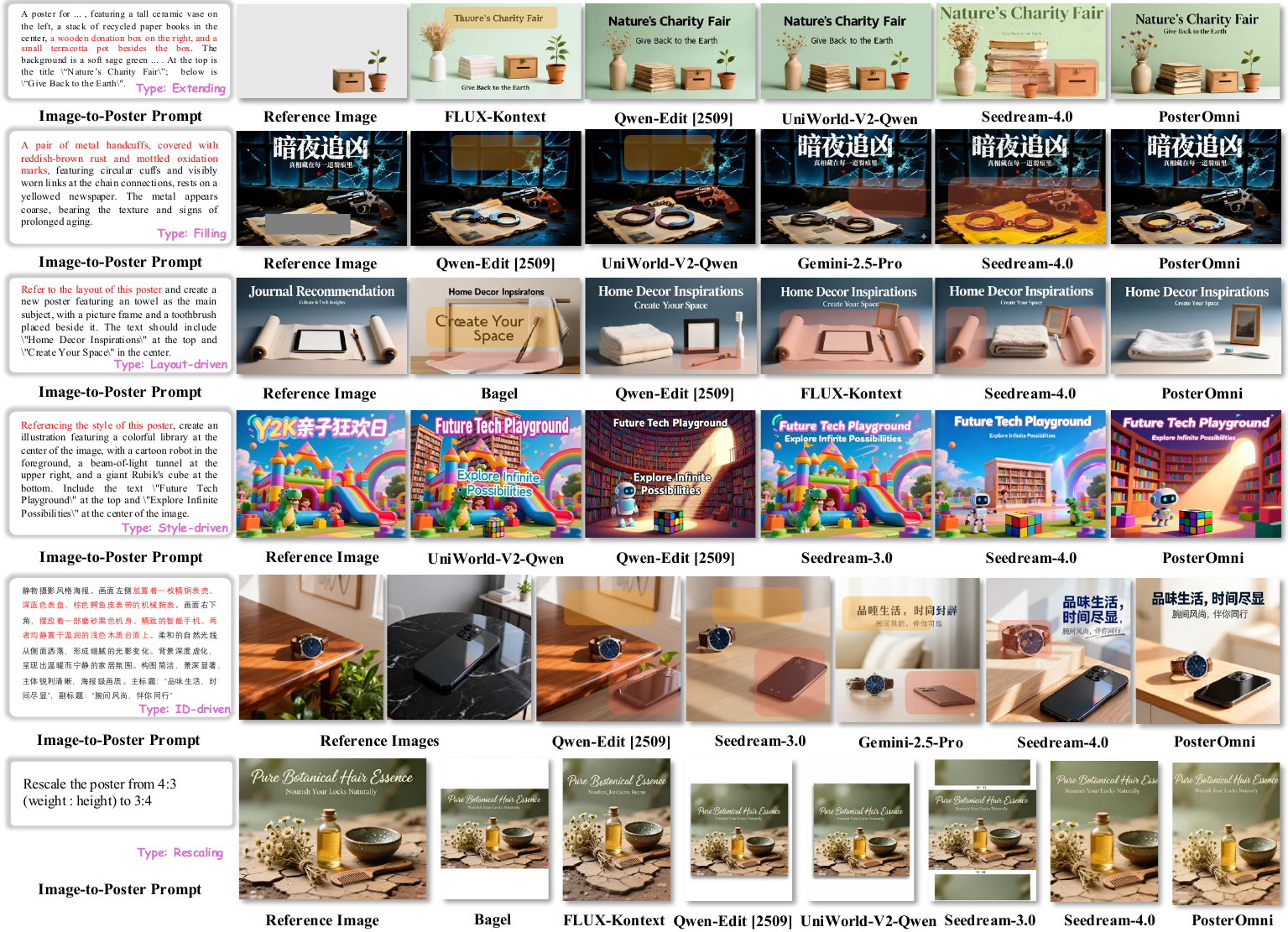}
    \caption{\textbf{Visual comparison} of different model outputs. \colorbox{lred}{Red boxes} highlight errors and distorted entities, while \colorbox{lyellow}{yellow boxes} indicate incorrect or missing text elements. Compared to other methods, our method is able to accomplish all image-generated poster tasks more effectively, while also achieving excellent aesthetic quality.}
    \label{fig:visual_1}
\end{figure*}
\subsection{Implementation}
For PosterOmni, we use Qwen-Image-Edit [2509]~\cite{qwenimage} as the base model. Local editing and global creation experts are trained with rank-128 LoRA using AdamW (lr = 1e-4) for 100K and 50K steps. During task distillation, the student adopts a half-rank LoRA (64), which we find sufficient for integrating expert knowledge without redundancy. The distillation weight is $\lambda_{\text{E}} = 1$, with lr = 2e-4, trained for 4000 steps. For the PosterOmni Reward Model, Qwen3-VL~\cite{yang2025qwen3} is fine-tuned with a rank-64 LoRA (lr = 1e-4) for 6000 steps. In the final Omni-Edit RL stage, only a lightweight rank-32 LoRA is updated on top of PosterOmni-SFT for 500 steps. All stages use AdamW~\cite{adamw} for stable convergence, and expert training samples are drawn randomly within each task category to maintain balance.

\subsection{PosterOmni-Bench}
As described in Sec.\ref{data construction}, we build PosterOmni-Bench using our automated data pipeline. The benchmark spans six tasks—extending, filling, rescaling, identity-driven, layout-driven, and style-driven generation. Specifically, it includes 540 Chinese prompts (PosterOmni-Bench-cn) and 480 English prompts (PosterOmni-Bench-en), evenly distributed across six poster themes with both single-image and multi-image cases. It serves to evaluate existing models’ image-to-poster capabilities.

We benchmark a wide range of models, including leading open-source editing models and commercial systems. Models supporting Chinese editing are tested on both PosterOmni-Bench-en and PosterOmni-Bench-cn, while English-only models are evaluated on PosterOmni-Bench-en.
Inspired by ImgEdit~\cite{ye2025imgedit}, we use Gemini-2.5-Pro for evaluation. As a strong VLM, it scores both general poster aesthetics and task completion on a 1–5 scale, and we use a weighted average as the final metric. Evaluation prompts and further details are provided in the supp..

\begin{table*}[t]
\centering
\caption{Quantitative comparison results on proposed PosterOmni-Bench. We use Gemini-2.5-Pro~\cite{gemini} for evaluation poster creation results. \textbf{Bold} indicates the best performance. We highlight the \colorbox{best}{best} and \colorbox{second}{second} metrics. The numbers before and after “/” correspond to the PosterOmni-Bench-en and PosterOmni-Bench-cn, respectively.} 
\setlength{\tabcolsep}{5pt}
\scalebox{0.68}{
\renewcommand{\arraystretch}{1.1}
\begin{tabular}{l|ccccccc}
\toprule[1.5pt]
\textbf{Model} & \textbf{Extending} & \textbf{Filling} & \textbf{Rescaling} & \textbf{Id-consis.} & \textbf{Layout-dri.} & \textbf{Style-dri.} & \textbf{Overall} $\uparrow$ \\
\midrule
ICEdit~\citep{icedit}~\textcolor{gray_venue}{\small{(Open)}} & 1.99 /\ --  & 3.21/\ --  & 1.73 /\ -- & 1.59 /\ -- & 1.53 /\ -- & 1.67 /\ -- & 1.95 /\ --  \\
Step1X-Edit~\citep{step1x-edit}~\textcolor{gray_venue}{\small{(Open)}} & 3.04 /\ 3.67 & 4.35 /\ 4.21 & 1.60 /\ 1.75 & 1.70 /\ 2.14 & 1.63 /\ 1.82 &1.57 /\ 1.79 & 2.31 /\ 2.56  \\
BAGEL~\citep{bagel}~\textcolor{gray_venue}{\small{(Open)}} & 2.33 /\ 2.84 & 2.77 /\ 2.67 & 1.77 /\ 1.40 & 1.92 /\ 2.29 & 2.34 /\ 3.03 & 1.85 /\ 2.34 & 2.15 /\ 2.43  \\
OmniGen2~\citep{omnigen2}~\textcolor{gray_venue}{\small{(Open)}} & 2.56 /\ -- & 2.32 /\ -- & 1.61 /\ -- & 3.25 /\ -- & 2.22 /\ -- & 1.84 /\ -- & 2.59 /\ --  \\
FLUX.1 Kontext [dev]~\citep{fluxkontext}~\textcolor{gray_venue}{\small{(Open)}}  & 3.12 /\ -- & 3.61 /\ -- &  3.16 /\ -- & 3.39 /\ -- & 3.03 /\ -- & 2.88 /\ -- & 3.20 /\ --  \\
Qwen-Image-Edit [2509]~\citep{qwenimage}~\textcolor{gray_venue}{\small{(Open)}} & 4.28 /\ 4.24 & 3.95 /\ 3.79 & 3.40 /\ 3.54& 3.06 /\ 3.37& 3.44 /\ 2.97& 2.91 /\ 2.83& 3.51 /\ 3.46 \\
UniWorld-V2-Qwen-Image-Edit~\cite{uniworldv2}~\textcolor{gray_venue}{\small{(Open)}} & 4.25 /\ 4.22 & 3.57 /\ 3.18 & 3.07 /\ 3.23& 2.87 /\ 3.20 & 3.66 /\ 3.79& 3.14 /\ 2.85 & 3.42 /\ 3.41 \\
\midrule 
Seedream-3.0~\citep{seedream3.0}~\textcolor{gray_venue}{\small{(Close)}} & 3.52 /\ 3.76 & 3.40 /\ 3.52 & 2.38 /\ 2.84 & 2.88 /\ 3.30 & 2.68 /\ 3.04 & 2.32 /\ 2.82 & 2.86 /\ 3.21 \\
Seedream-4.0~\citep{seedream4.0}~\textcolor{gray_venue}{\small{(Close)}} & \colorbox{second}{4.41} /\ \colorbox{second}{4.57} &  \colorbox{second}{4.44} /\ \colorbox{second}{4.64}& \colorbox{best}{4.00} /\ \colorbox{second}{3.69}& \colorbox{best}{4.53} /\ \colorbox{best}{4.62} & \colorbox{second}{4.05} /\ \colorbox{second}{4.22} & \colorbox{best}{4.23} /\ \colorbox{second}{4.31} &  \colorbox{best}{4.28} /\ \colorbox{second}{4.34} \\
\midrule
PosterOmni (Ours)& \colorbox{best}{4.76} /\ \colorbox{best}{4.72} & \colorbox{best}{4.69} /\ \colorbox{best}{4.77} & \colorbox{second}{3.97} /\ \colorbox{best}{3.81} & \colorbox{second}{3.98} /\ \colorbox{second}{4.23} & \colorbox{best}{4.20} /\ \colorbox{best}{4.35} & \colorbox{second}{3.99} /\ \colorbox{best}{4.36} & \colorbox{second}{4.27} /\ \colorbox{best}{4.37} \\
\rowcolor{green!5}
\textit{vs. Baseline (Qwen-Image-Edit [2509])} &
\textcolor{gray}{\textbf{+0.48 /\ +0.48}} &
\textcolor{gray}{\textbf{+0.74 /\ +0.98}} &
\textcolor{gray}{\textbf{+0.57 /\ +0.27}} &
\textcolor{gray}{\textbf{+0.92 /\ +0.86}} &
\textcolor{gray}{\textbf{+0.76 /\ +1.38}} &
\textcolor{gray}{\textbf{+1.08 /\ +1.53}} &
\textcolor{gray}{\textbf{+0.76 /\ +0.91}} \\
\bottomrule[1.5pt]
\end{tabular}}
\label{tab_posteromni-bench} 
\end{table*}

\subsection{Quantitative Results and Comparisons}
Table \ref{tab_posteromni-bench} summarizes the results on PosterOmni-Bench. Overall, PosterOmni delivers clear improvements across all six tasks. On the local editing tasks—including extending, filling, rescaling, and ID-driven generation—the model outperforms the Qwen-Image-Edit [2509] baseline~\cite{qwenimage} by a noticeable margin, with gains ranging from +0.48 to +0.98.
For the two global creation tasks, layout-driven and style-driven poster generation, PosterOmni also shows substantial advantages over the base model and all other open-source systems. When considering all six tasks together, our performance comes close to and even exceeds the latest proprietary models, such as Seedream-4.0~\cite{seedream4.0}, highlighting the practical value of our approach.
These improvements reflect the effectiveness of our task-distillation SFT and unified reward feedback, which help the model follow image-to-poster instructions while producing more coherent and aesthetically consistent results. Taken together, the results indicate that our unified poster creation model can successfully handle both local editing and higher-level poster creation without relying on separate expert models.
\subsection{Qualitative Results and Comparisons}
Figure~\ref{fig:visual_1} presents visual comparisons across all six poster creation tasks, covering both open-source baselines and strong commercial systems such as Seedream-3.0~\cite{seedream3.0}, Seedream-4.0~\cite{seedream4.0}, and Gemini-2.5-Pro~\cite{gemini}. Across the local editing tasks, other models often generate incorrect entities, incomplete regions, or missing text—highlighted by the red and yellow boxes. PosterOmni preserves structure and semantics more reliably, producing edits that blend naturally with the reference image and maintain clean, readable typography. For global creation tasks, including layout-driven and style-driven generation, PosterOmni also shows clearer layout logic and more consistent aesthetics. Several commercial models optimized for general image editing tend to copy the reference image directly or struggle to coordinate multiple elements. PosterOmni follows instructions more faithfully, generating posters with coherent themes, balanced text placement, and stronger overall composition. These qualitative results show that PosterOmni effectively handles both precise local edits and higher-level poster design, delivering outputs competitive with advanced proprietary systems.

\section{Ablation Study}
To evaluate the contribution of each core component in the PosterOmni framework, we conduct a comprehensive ablation study on the PosterOmni-Bench-en. We select representative local editing (extend) and global creation (layout-driven) tasks to analyze the impact of each module on editing precision, aesthetic quality, and holistic consistency. For fair comparison, all experimental settings and parameters are kept identical to the main experiments.

\begin{table}[t]
    \centering
    \caption{Ablation study of our task distillation. Scores are averaged
on the selected local (extend) and global (layout) tasks.}
    \label{tab:ab_task_distill}
    \scalebox{0.85}{
        \begin{tabular}{l|c}
            \toprule[1.5pt]
            \textbf{Model} & \textbf{PosterOmni (L / G)$\uparrow$} \\
            \midrule
            \textbf{Qwen-Image-Edit [2509]} & 4.28 /\ 3.44 \\
            \midrule
            (i). Mixed Training (L + G) &  4.33 /\ 3.72 \\
            (ii). + Task-specific Expert (Local) & \colorbox{best}{4.48} /\ 2.79 \\
            (iii). + Task-specific Expert (Global) & 3.35 /\ \colorbox{best}{3.96} \\
            (iv). + Task Distillation &  4.39 /\ 3.82  \\
            (v). (iv) + Aux. Loss (PosterOmni-SFT) & \colorbox{second}{4.43} /\ \colorbox{second}{3.89} \\
            \bottomrule[1.5pt]
        \end{tabular}
    }
\end{table}

\noindent{\textbf{\textit{Effectiveness of Task Distillation:}}}
We compare our task distillation strategy with the base model and four varients: (i) joint training on all tasks; (ii) individually trained local experts; (iii) individually trained global experts; and (iv) distillation without auxiliary text-rendering loss.
As shown in Table \ref{tab:ab_task_distill}, the base model exhibits limited cross-task generalization, while mixed training still suffers from interference between low-level editing and high-level compositional objectives. Individual experts perform well only on their own tasks but fail to transfer to others. Removing the auxiliary loss weakens text clarity.
In contrast, our distilled model maintains strong performance across both local and global tasks, achieving expert-level precision while preserving compositional quality.

\begin{table}[ht]
    \centering
    \caption{Ablation of unified reward feedback. Scores are averaged on the selected local (extend) and global (layout) tasks.}
    \label{tab:ab_reward_feedback}
    \scalebox{0.85}{
    \begin{tabular}{l|c}
            \toprule[1.5pt]
            \textbf{Model} & \textbf{PosterOmni  (L / G)$ \uparrow$} \\
            \midrule
            \textbf{PosterOmni-SFT} & 4.43 /\ 3.89 \\
            \midrule
        (i) + VLM-based $R_{\text{v}}$~\cite{uniworldv2} + Omni-Edit RL & 4.58 /\ 3.97  \\
        (ii) + Unified $R_{\text{omni}}$ + FlowGRPO~\cite{liu2025flowgrpo} & \colorbox{second}{4.65} /\  \colorbox{second}{4.08} \\
        (iii) + Unified $R_{\text{omni}}$ + Omni-Edit RL (Ours)  & \colorbox{best}{4.76} /\ \colorbox{best}{4.20}  \\
        \bottomrule[1.5pt]
    \end{tabular}}
\end{table}

\noindent{\textbf{\textit{Effectiveness of Unified Reward Feedback:}}}
We further investigate the effectiveness of unified reward feedback, encompassing both reward model training and Omni-Edit reinforcement learning. As shown in Table~\ref{tab:ab_reward_feedback}, removing the reward model leads to weaker aesthetic alignment and less stable layout coherence. We also compare against two alternative strategies: the UniWorld-V2~\cite{uniworldv2} setting, which scales state-of-the-art VLMs~\cite{yang2025qwen3} as the reward model, and FlowGRPO~\cite{liu2025flowgrpo}, which performs gradient-based policy optimization. In contrast, integrating our unified $R_{\text{omni}}$ with Omni-Edit RL yields consistently higher scores on both local and global tasks, demonstrating that unified reward feedback provides coherent guidance through balanced task-specific and general optimization signals. More abl. studies can be found in supp..

\section{Conclusion}
We presented PosterOmni, a generalized model for image-to-poster creation that brings together local editing and global design within a single framework. With task-distillation SFT and unified reward feedback, the model learns both precise visual adjustments and coherent poster-level composition. Experiments on PosterOmni-Bench show clear improvements across all tasks, surpassing all open-source systems and approaching the quality of leading commercial models. Overall, PosterOmni demonstrates that a unified model can effectively handle the diverse requirements of real-world poster generation. 
\clearpage

This is supplementary material for \emph{PosterOmni: Generalized Artistic Poster Creation via Task Distillation and Unified Reward Feedback}. 

We present the following materials in this supplementary document:

\begin{itemize}
    \item \textbf{Sec.~\ref{sec:data construction}} Details of our PosterOmni data suite (PosterOmni-200K and PosterOmni-Bench), covering prompt design, multimodal filtering, task-specific image-to-poster construction pipelines, and keyword/topic coverage.
    \item \textbf{Sec.~\ref{sec:rm training}} Construction of the PosterOmni reward training dataset and implementation details of the unified PosterOmni reward model $R_{\text{omni}}$.
    \item \textbf{Sec.~\ref{sec:user-study}} User study setup and results, including the human evaluation protocol and win/tie/loss statistics against open-source and proprietary baselines.
    \item \textbf{Sec.~\ref{sec:addition abl}} Additional ablation studies on reward model design and expert integration strategies.
    \item \textbf{Sec.~\ref{sec:addition vc}} Additional visual comparisons across all six image-to-poster tasks, illustrating qualitative differences between PosterOmni and competing methods.
    \item \textbf{Sec.~\ref{sec:future work}} Limitations and future work of the PosterOmni.
\end{itemize}

\section{Details of {PosterOmni} Data (PosterOmni-200K and PosterOmni-Bench)}\label{sec:data construction}

In this section, we provide additional details of our data suite PosterOmni data, which consists of the training set \textbf{PosterOmni-200K} and the evaluation benchmark \textbf{PosterOmni-Bench}. The main paper briefly introduces the automated pipeline in Sec.~3.1 of our manscript; here we elaborate on task-specific construction, multimodal filtering, and topic coverage.

\subsection{Prompt Design and Base Text-to-Image Corpus}

We first build a diverse text-to-image corpus that mimics real poster design scenarios. Following the meta-prompt in Fig.~\ref{prompt:txt2img-construction}, we sample a \emph{category} (e.g., products, food, events/travel, education, nature, entertainment), a \emph{scenario} (e.g., ``family feast'', ``AI summit''), and a \emph{style tag} (e.g., Swiss grid, watercolor). For each triplet, a VLM (GPT~\cite{openai_gpt5_en}/Qwen3~\cite{yang2025qwen3}) plays the role of a ``creative director'' and writes a fluent image-to-poster prompt specifying (1) main subjects, (2) spatial composition, (3) overall mood and color palette, and (4) 1–3 pieces of rendered text with approximate positions (title, slogan, time/place).

We instantiate the template in both English and Chinese, leading to bilingual prompts with consistent semantics. Multiple candidate images are then generated per prompt by strong text-to-image models (Qwen-Image~\cite{qwenimage} and FLUX-style generators), which form the base images from which all downstream tasks are constructed.

\begin{figure*}[!t]
    \centering
    \includegraphics[width=16.5cm]{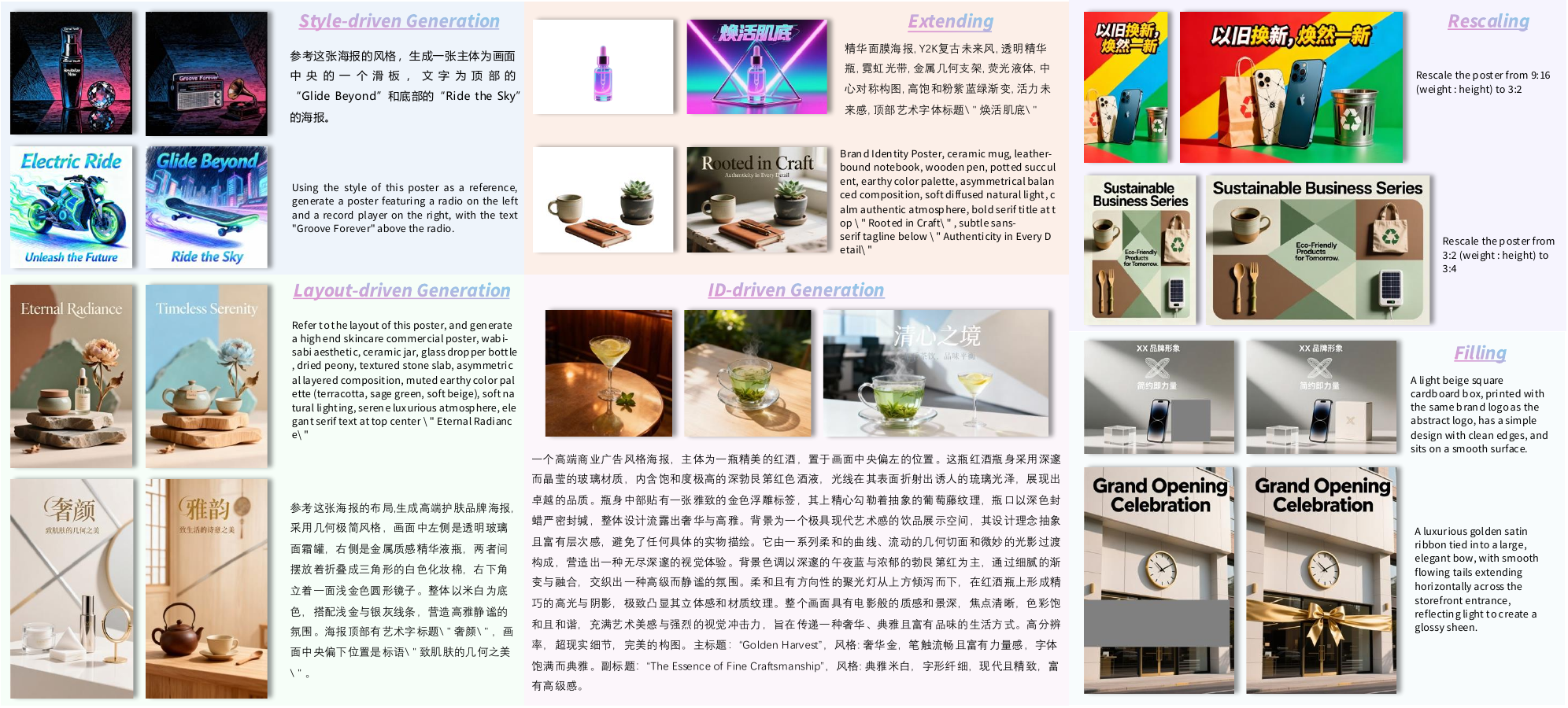}
    \caption{\textbf{Examples from our PosterOmni-data ( PosterOmni-200K and PosterOmni-Bench).} For each of the six core image-to-poster tasks—style-driven generation, layout-driven generation, ID-driven generation, extending, rescaling, and filling—we show the reference image(s) together with the corresponding image-to-poster prompts in both English and Chinese. The examples illustrate diverse commercial scenarios, layouts, and visual styles, as well as the explicit task-specific instructions.}
    \label{example_of our posteromni data}
\end{figure*}

\subsection{Multimodal Filtering for {PosterOmni}-200K and {PosterOmni}-Bench}

To ensure that the synthetic posters are usable for supervised training and reliable evaluation, we apply a multi-stage multimodal filtering pipeline (Fig.~2 in our manscript).

\noindent{\textbf{Training set (PosterOmni-200K).}}
For each (prompt, image) pair we perform:
\begin{itemize}
    \item \textbf{OCR and text sanity checks.} PaddleOCR~\cite{cui2025paddleocr} is used to extract rendered text; we reject images where the decoded strings are unreadable or deviate too much from the prompt keywords (e.g., wrong language, heavy corruption).
    \item \textbf{Vision--language consistency.} We embed both the prompt and image with Jina-clip-v2~\cite{jina-clip-v2} and drop samples whose similarity falls below a threshold, which removes cases where the layout or subject semantics are obviously mismatched with the description.
    \item \textbf{Layout and clutter constraints.} Simple heuristics on text box count, text area ratio, and foreground–background separation filter out extremely cluttered or almost text-free images, so that each sample still resembles a reasonable poster layout.
\end{itemize}

\begin{wrapfigure}{r}{8.5cm}
    \centering
    \includegraphics[width=8.5cm]{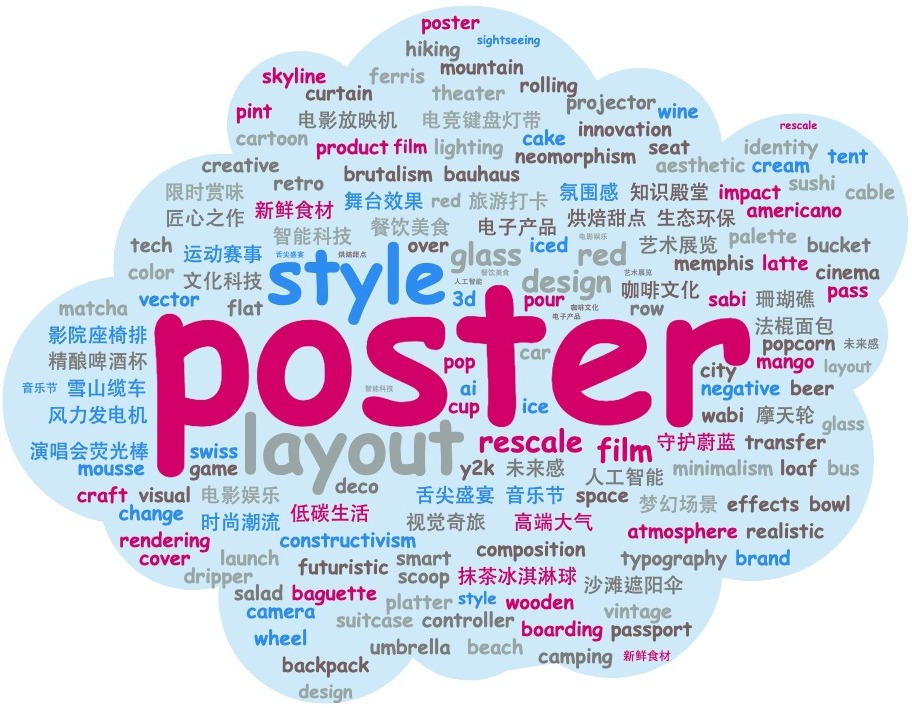}
    \caption{\textbf{Word cloud of representative keywords in PosterOmni-data.}  It illustrates the diversity of our dataset in terms of styles, layouts, contents and themes, covering a broad range of real-world poster scenarios in both English and Chinese.}
    \label{cloudword}
\end{wrapfigure}
Only pairs passing all checks are used as sources for building task-specific image-to-poster pairs.

\noindent{\textbf{Benchmark (PosterOmni-Bench).}}
For the test benchmark, we adopt stricter filtering:
\begin{itemize}
    \item \textbf{Task suitability via VLM.} Given a candidate image, Gemini-2.5-Flash~\cite{gemini} is queried with the task-matching meta-prompt in Fig.~\ref{prompt:task-matching}. It must assign exactly one label from \{\texttt{EXTENDING}, \texttt{FILLING}, \texttt{RESCALING}, \texttt{ID-DRIVEN}, \texttt{LAYOUT-DRIVEN}, \texttt{STYLE-DRIVEN}, \texttt{NONE}\}. We keep only samples that receive a confident, non-\texttt{NONE} label consistent with our intended task.
    \item \textbf{Manual spot-checking.} For each task and theme, we manually review a subset of images to verify that the predicted task matches human intuition (e.g., that an ``extending'' candidate indeed has expandable background, that a ``layout'' case exposes a clear grid/compositional structure).

\end{itemize}
This procedure yields 480 English and 540 Chinese prompts paired with reference images, balanced across six themes and six tasks, as described in the main paper.

\begin{figure*}[!t]
    \centering
    \setlength{\belowcaptionskip}{-0cm}
    \includegraphics[width=16.5cm]{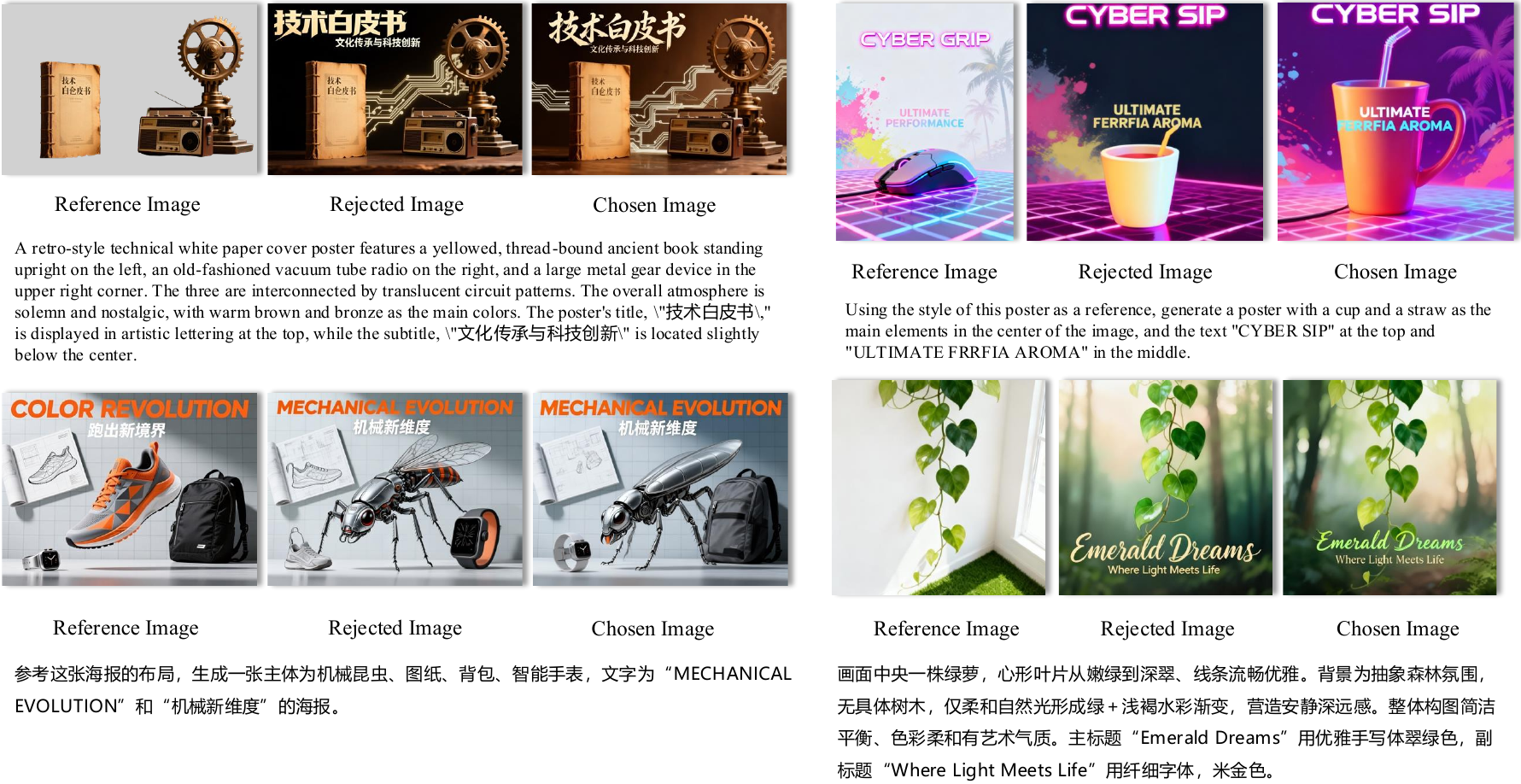}
    \caption{\textbf{Examples of preference pairs for PosterOmni Reward Training.}
    For several representative style-driven and layout-driven cases, we show the
    reference image together with the rejected and chosen candidates produced
    by PosterOmni-SFT, as well as the corresponding image-to-poster prompts in
    English and Chinese. Each triplet (reference, rejected, chosen) constitutes
    a concrete example of the preference pairs used to train the unified reward
    model $R_{\text{omni}}$.}
    \label{fig:reward_training_data}
\end{figure*}

\subsection{Task-Specific Image-to-Poster Construction}

Starting from the filtered text-to-image corpus, we construct paired image-to-poster samples for six tasks by applying modular, task-specific transformations (also summarized in Fig.~2 of our manscript). Examples of the resulting input–output pairs are shown in Fig.~\ref{example_of our posteromni data}.

\noindent{\textbf{Extending.}}
Given a reference poster, SAM-2~\cite{sam2} is used to segment the main subject and foreground elements. We dilate the foreground mask and treat the remaining area as ``extendable'' background. The input image is obtained by cropping the canvas around the subject, while the target poster retains the original full canvas. The image-to-poster prompt explicitly asks the model to extend the canvas, this encourages learning seamless background extension and composition completion.

\noindent{\textbf{Filling.}}
For the filling task, we first sample one or more localized regions (such as a logo slot, product placeholder, or empty billboard) using SAM-2~\cite{sam2} masks and simple geometric rules. The input is created by erasing these regions to obtain a hole image; the target is the original poster. Fill prompts ask the model to regenerate appropriate content inside the hole (e.g., ``replace the empty stand with a perfume bottle'') under the same style and lighting.

\noindent{\textbf{Rescaling.}}
Rescaling pairs simulate aspect-ratio changes without distorting the main subjects. For a clean poster, we use the SOTA commercial models/crop methods to change the central region to a different aspect ratio (e.g., from $2\!:\!3$ to $4\!:\!3$) and use BrushNet~\cite{ju2024brushnet} to extend the margins where necessary so that the central content stays intact. The cropped or partially extended image is used as the input, while the fully adjusted poster serves as the output. Prompts describe the target ratio (as in PosterOmni-Bench) to teach the model aspect-ratio-aware composition.

\noindent{\textbf{Identity-driven generation.}}
We treat the original poster as an ID reference and generate a new scene featuring the same key subject. PaddleDet~\cite{paddledet} first detects identity-critical objects (e.g., a specific drink can, branded product, or mascot). We then use SOTA models to synthesize new image(s) where the subject appears in a different pose or environment but with consistent fine-grained identity (shape, color pattern, logo). The input consists of the generated image(s), and the output is the reference poster, supervised by prompts that stress preserving identity while changing context.

\noindent{\textbf{Layout-driven generation.}}
Here the input is a clean layout template with recognizable blocks (hero image area, text zones, logo strip, etc.). We use VLMs (e.g., Gemini-2.5-Pro~\cite{gemini}) or simple heuristic rules to extract a coarse layout graph, then ask the SOTA models to ``follow the layout'' but replace the content (e.g., new products and background). Then, SOTA VLM feedback is used to construct image-to-poster prompts to form a complete data pair.

\noindent{\textbf{Style-driven generation.}}
For style-driven generation, the construction is analogous but focuses on visual treatment rather than spatial structure. Given a reference poster, we treat it as a style template and use VLMs to summarize key stylistic attributes such as color palette, rendering texture, lighting, and typography (e.g., ``vaporwave cyberpunk''). We then require the existing editing model to replace parts of the text and objects in the scene while preserving reasonable consistency with the main stylistic features and scene semantics. The VLM is then used again to generate corresponding image-to-poster conversion prompts. Therefore, the input reference poster and target poster are stylistically similar but differ in specific content.

\subsection{Keyword Distribution and Topic Coverage}
To better visualize the semantic coverage of \textsc{PosterOmni}-data, Fig.~\ref{cloudword} shows a word cloud built from all English and Chinese prompts. Large keywords such as ``poster'', ``layout'', ``style'', ``rescale'', and ``film'' correspond to our core tasks and typical poster scenarios, while medium and small words cover product categories (e.g., coffee, skincare, camera), event types (e.g., concert, marathon, exhibition), and design attributes (e.g., ``minimalism'', ``memphis'', ``cream tone''). The mixture of bilingual tokens indicates that the dataset spans both Chinese and English markets and emphasizes realistic commercial usage rather than toy scenes. In addition to the high-quality data generated by our pipeline, \textsc{PosterOmni}-data also includes a small portion ($<10\%$) of in-house poster data; these samples are processed with the same processes.

\section{PosterOmni Reward Training Dataset and Model Details}\label{sec:rm training}

\subsection{PosterOmni Reward Dataset Construction}
To clarify the data used for training the unified PosterOmni Reward Model
$R_{\text{omni}}$, we summarize the construction pipeline and basic statistics
here. As illustrated in Fig.~\ref{fig:reward_training_data}, starting from the
SFT-trained PosterOmni model, we generate candidate posters for all six
image-to-poster task types. Candidate images are grouped into pairs, each pair
sharing the same input context and task description. We then query
Gemini-2.5-Pro~\cite{gemini} with the preference prompt shown in
Fig.~\ref{prompt:5-3-preference} to obtain an automatic choice between the two
candidates. Pairs for which Gemini indicates a clear preference and at least
one candidate already satisfies basic poster quality are kept, while pairs
where both candidates are obviously broken (e.g., unreadable text, collapsed
layout) are discarded. This step acts as a coarse filter and provides an
initial ranking signal.

On the remaining pairs, human annotators perform a light review using the same
task-specific criteria as in Fig.~\ref{prompt:5-3-preference}, correcting
Gemini's decisions when necessary and discarding ambiguous or noisy cases.
After this two-stage filtering and review, we obtain roughly $60$K clean
preference pairs across all tasks. The distribution over task types is
slightly imbalanced but covers both local editing (rescale, fill, extend,
ID-driven) and global creation (layout-driven, style-driven) cases. For reward
training, each labeled pair $(I_{\text{chosen}}, I_{\text{rejected}})$
additionally yields a simple negative pair by treating the original input
image $I_{\text{in}}$ as the less preferred sample and $I_{\text{chosen}}$ as
the preferred one, so that $R_{\text{omni}}$ learns to favor complete
poster-like edits over raw inputs. Tab.~\ref{tab:supp_reward_stats} reports
the approximate per-task statistics used in our experiments.

\begin{table}[t]
\centering
\caption{Approximate statistics of the PosterOmni preference dataset used to train 
$R_{\text{omni}}$. Each row reports the number of human-checked preference pairs for 
a task type. During reward training, we further augment the data with input--output 
negative pairs $(I_{\text{in}}, I_{\text{chosen}})$; the last column shows the 
approximate fraction of such extra negatives among all comparisons.}
\label{tab:supp_reward_stats}
\scalebox{0.85}{
\begin{tabular}{lccc}
\toprule
\textbf{Task type} & \textbf{\#Preference pairs} & \textbf{Share of all pairs} & \textbf{Extra negative-pair ratio} \\
\midrule
Poster Rescale              & 11{,}000 & $\approx 18\%$ & $33.3\%\,$ (1:3) \\
Poster Fill                 &  9{,}000 & $\approx 15\%$ & $33.3\%\,$ (1:3) \\
Poster Extend               & 10{,}000 & $\approx 17\%$ & $33.3\%\,$ (1:3) \\
Identity-driven   &  8{,}000 & $\approx 13\%$ & $33.3\%\,$ (1:3) \\
Layout-driven     & 11{,}000 & $\approx 18\%$ & $33.3\%\,$ (1:3) \\
Style-driven      & 11{,}000 & $\approx 18\%$ & $33.3\%\,$ (1:3) \\
\midrule
\textbf{Overall}            & 60{,}000 & $100\%$        & $33.3\%\,$ (1:3) \\
\bottomrule
\end{tabular}
}
\end{table}

\subsection{PosterOmni Reward Model Architecture and Training}

Based on the preference data described above, we instantiate $R_{\text{omni}}$ on top of the Qwen3VL~\cite{yang2025qwen3} encoder with a lightweight regression head. For each quadruplet $(I_{\text{in}}, p_t, \text{edit}, I)$, we treat $I$ as the candidate poster to be scored. The image $I$ is fed to the vision branch of Qwen3VL, while the text branch concatenates the task prompt $p_t$, the editing description \texttt{edit}, and a short task-type tag (e.g., ``[Task: Layout-driven generation]''). We take the pooled multimodal representation and pass it through a small MLP head to obtain a scalar reward $r_\theta(I) \in \mathbb{R}$. Since each task is accompanied by explicit instructions and a task-type indicator, the reward model learns to distinguish fine-grained quality differences between candidates under the \emph{same} task while sharing parameters across \emph{different} tasks. In this way, preference learning mainly depends on relative scores within each task, yet results in a single unified PosterOmni reward model applicable to all image-to-poster settings.

\begin{figure*}[!t]
    \centering
    \includegraphics[width=16.5cm]{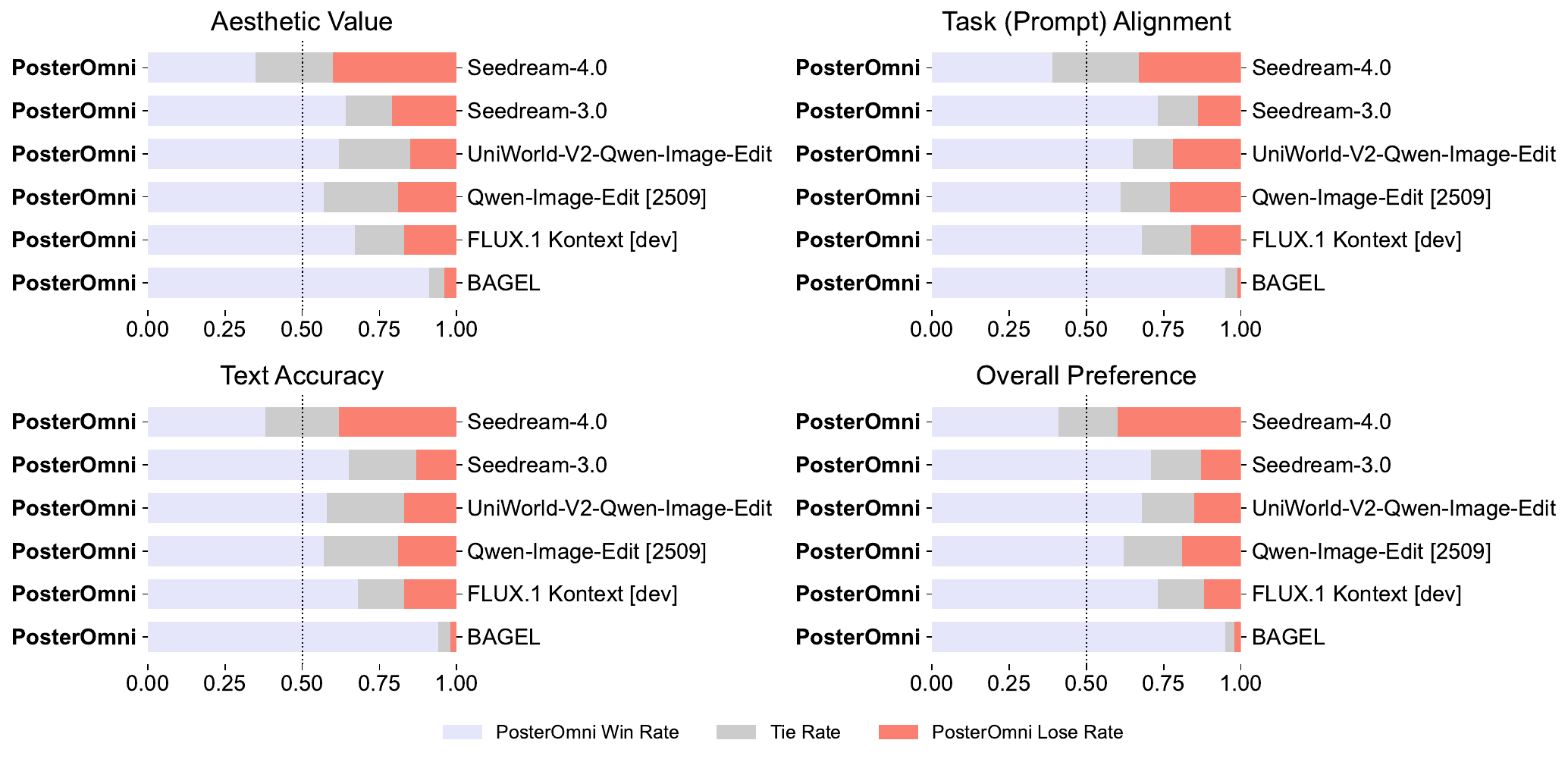}
    \caption{\textbf{Human preference study for image-to-poster generation.}
    We compare PosterOmni with six competing systems (Seedream-4.0~\cite{seedream4.0}, Seedream-3.0~\cite{seedream3.0},
    UniWorld-V2–Qwen-Image-Edit~\cite{uniworldv2}, Qwen-Image-Edit~[2509]~\cite{qwenimage}, FLUX.1 Kontext [dev]~\cite{fluxkontext}, and BAGEL~\cite{bagel})
    under four criteria: Aesthetic Value, Task (Prompt) Alignment, Text Accuracy, and Overall Preference.
    For each pairwise comparison, bars report the fraction of cases in which PosterOmni is preferred
    (light purple), tied (gray), or worse (red) than the competing model.
    The vertical dashed line at 0.5 denotes parity; bars extending to the right indicate that PosterOmni is more often favored than the corresponding baseline. Overall, PosterOmni significantly outperforms all existing open-source models and performs on par with the state-of-the-art proprietary system Seedream-4.0.}
    \label{comparison_plot}
\end{figure*}

\section{User Study}\label{sec:user-study}

Besides the automatic metrics reported the in mainscript, we further conduct a human
preference study to directly assess the perceptual quality of different image-to-poster generation
systems. Our goal is to measure how often PosterOmni is preferred by human users compared with
both open-source and proprietary baselines.

\noindent{\textbf{Setup.}}
We randomly sample 150 prompts from PosterOmni-Bench-en (in order to compare all models), covering all six poster-editing tasks
(extend, fill, rescale, ID-driven, layout-driven, and style-driven generation).
For each prompt, we generate posters using PosterOmni and six competing systems
(Seedream-4.0~\cite{seedream4.0}, Seedream-3.0~\cite{seedream3.0}, UniWorld-V2--Qwen-Image-Edit~\cite{uniworldv2}, Qwen-Image-Edit~[2509]~\cite{qwenimage},
FLUX.1 Kontext [dev]~\cite{fluxkontext}, and BAGEL~\cite{bagel}).
We recruit six experienced poster designers, all of whom have at least two years of professional design experience. Each rater is presented with pairwise comparisons between PosterOmni and one baseline at a time, under a randomized order of prompts and model sides (left/right) to avoid bias.

\noindent{\textbf{Protocol and metrics.}}
For each comparison, raters are asked to judge the two posters along four criteria:
(i) \emph{Aesthetic Value} (overall visual appeal and layout harmony),
(ii) \emph{Task (Prompt) Alignment} (whether the poster correctly follows the editing instruction
and preserves required content/layout),
(iii) \emph{Text Accuracy} (correctness and legibility of rendered text), and
(iv) \emph{Overall Preference} (which poster they would choose to use in a real project).
For every criterion, raters choose one of three options:
``PosterOmni is better'', ``Tie'', or ``Baseline is better''.
Given all annotations, we compute for each baseline and criterion the
\emph{win rate} $w$ (fraction of comparisons where PosterOmni is preferred),
\emph{tie rate} $t$, and \emph{loss rate} $\ell$ (fraction where the baseline is preferred),
normalized so that $w + t + \ell = 1$.
These win/tie/loss rates are reported in Fig.~\ref{comparison_plot}.

\noindent{\textbf{Results.}}
As shown in Fig.~\ref{comparison_plot}, PosterOmni achieves consistently higher win rates than all existing open-source systems across all four criteria, with especially strong gains in Task (Prompt) Alignment.
Against the state-of-the-art proprietary system Seedream-4.0, PosterOmni attains comparable performance: their win/loss bars are close to the $0.5$ parity line for all criteria, indicating that users find the two systems essentially on par.
Overall, the user study confirms that PosterOmni not only improves objective metrics, but also delivers posters that human designers genuinely prefer in real design scenarios.

\section{Additional Ablation Studies.}\label{sec:addition abl}
\subsection{Ablation on PosterOmni Reward Model Design}

In this section, we supplement the ablation experiments on PosterOmni by
focusing on how the design of the unified reward model $R_{\text{omni}}$
affects downstream image-to-poster quality. For each variant of
$R_{\text{omni}}$, we keep the Omni-Edit RL procedure (DiffusionNFT-based
policy optimization) and all hyper-parameters fixed, and only swap the reward
model used to score generated samples. The final scores therefore reflect the
quality of the reward signal rather than changes in the RL algorithm.

Concretely, starting from the same preference pairs, we compare three designs:

\begin{itemize}
    \item \textbf{w/o Negative pairs}: we remove the additional input--output
    pairs $(I_{\text{in}}, I_{\text{chosen}})$ and train the reward model only
    on candidate--candidate preferences. In this case $R_{\text{omni}}$ mostly
    learns relative aesthetics between edited posters, without being explicitly
    penalized for staying too close to the raw input image.
    \item \textbf{w/o Image-to-poster prompt}: we keep all pairs but drop the
    full image-to-poster prompt from the text input of $R_{\text{omni}}$,
    leaving only the task-type tag (e.g., ``[Task: Layout]''). This variant
    emphasizes generic aesthetic preferences within each task, while largely
    ignoring the detailed creative brief and task-specific requirements.
    \item \textbf{Full $R_{\text{omni}}$ (Ours)}: the reward model uses both
    candidate--candidate and input--output pairs, and is conditioned on the
    complete image-to-poster prompt together with the task-type tag, forming a
    unified, instruction-aware reward across all tasks.
\end{itemize}

We evaluate these variants by applying the same Omni-Edit RL pipeline and
reporting the averaged scores on a local task (extend) and a global task
(layout-driven). As shown in Tab.~\ref{tab:ab_reward_model}, removing negative pairs leads to a clear
drop, especially on the global layout task. Compared with typical
text-to-image settings or cross-model comparisons, the image-to-poster
candidates produced by PosterOmni-SFT under the same instruction are already
relatively close to each other, so the quality gap within each pair can be
subtle. The additional negative pairs, constructed from the raw input image
and its output poster, provide clear, easy-to-recognize negative examples
and help $R_{\text{omni}}$ better learn what should be treated as a bad
output. Dropping the
image-to-poster prompt yields consistent degradation: the reward
model becomes biased toward purely aesthetic signals and tends to overlook
instruction-following for image-to-poster generation. The full unified
$R_{\text{omni}}$, trained with both negative pairs and prompt conditioning,
achieves the best balance on both local and global tasks.

\begin{table}[t]
    \centering
    \caption{Ablation study of PosterOmni Reward Model design. Scores are
    averaged on a local task (extend, L) and a global task (layout-driven, G).}
    \label{tab:ab_reward_model}
    \scalebox{0.85}{
        \begin{tabular}{l|c}
            \toprule[1.5pt]
            \textbf{Reward Model} & \textbf{PosterOmni (L / G)$\uparrow$} \\
            \midrule
            \textbf{PosterOmni-SFT (no RL) }       & 4.43 /\ 3.89 \\
            \midrule
            (i). w/o Negative pairs       & 4.64 /\ 4.03 \\
            (ii). w/o Image-to-poster prompt & \colorbox{second}{4.67} /\ \colorbox{second}{4.09} \\
            (iii). Full $R_{\text{omni}}$ (Ours) & \colorbox{best}{4.76} /\ \colorbox{best}{4.20} \\
            \bottomrule[1.5pt]
        \end{tabular}
    }
\end{table}

\textit{Additionally, our focus in this work is to develop an end-to-end PosterOmni framework, where
$R_{\text{omni}}$ is used as an internal optimization module for the
image-to-poster generator rather than as a stand-alone benchmarked model.
Consequently, we do not compare $R_{\text{omni}}$ against a wide range of
existing reward models. To the best of our knowledge, there is no reward model
specifically designed for image-to-poster generation, and our preference data
are tightly coupled with the PosterOmni-SFT generator and its task-specific
instructions. This mismatch in both task definition and data distribution
makes it difficult to fairly plug generic text-to-image or generic editing
reward models into our pipeline as drop-in replacements. We therefore restrict
our analysis to ablations on the design of $R_{\text{omni}}$ itself and
evaluate its quality indirectly through the final performance of PosterOmni,
leaving a more systematic study of cross-task reward transfer and reward-model
benchmarking to future work.}

\subsection{Ablation on Expert Integration Strategies}

Beyond the reward model design, we also study how to best integrate the
local- and global-editing experts into a single poster editor. Starting from
the task-specific experts $E_{\text{local}}$ and $E_{\text{global}}$ trained in
Sec.~3.2, we compare several ways of combining them into one model while
keeping the backbone and training budget fixed.

\begin{itemize}
    \item \textbf{Linear LoRA merge}: we directly interpolate the LoRA
    parameters of $E_{\text{local}}$ and $E_{\text{global}}$ with different
    weighting coefficients $\alpha \in \{0.25, 0.5, 0.75\}$, i.e.,
    $\Delta W = \alpha \Delta W_{\text{local}} + (1-\alpha)\Delta W_{\text{global}}$.
    This parameter-level fusion requires no extra training but ignores the large
    distribution gap between fine-grained local editing and global composition.
    \item \textbf{ZipLoRA fusion}: following ZipLoRA~\cite{shah2024ziplora},
    we compress and merge the two LoRA adapters into a single larger
    adapter. This variant explicitly reduces redundancy between experts,
    but still performs fusion purely in parameter space.
    \item \textbf{Task distillation (PosterOmni-SFT)}: our final design uses
    the two experts as teachers and trains a student editor with the
    distillation loss in Eq.~(2), jointly supervising the student on all six
    tasks with auxiliary text-rendering. This yields the unified
    PosterOmni-SFT model used in the main paper.
\end{itemize}

We evaluate these integration strategies on PosterOmni-Bench-en in the main ablation
study, and report Gemini scores in Tab.~\ref{tab:ab_expert_integration}.
Across all interpolation weights, linear merging leads to a clear degradation
on both tasks. In practice, we observe severe failure cases such as directly
copying the reference image, collapsing to a single dominant expert, or
producing nearly identical outputs for different task types, which is
unacceptable for a multi-task poster editor. ZipLoRA fusion provides a
slightly better balance, but still suffers from task interference and
distorted layouts: fusing heterogeneous experts only in parameter space
cannot preserve their complementary behaviours when the task set is diverse.
In contrast, the task-distilled PosterOmni-SFT consistently achieves the best
scores, showing that learning from expert outputs is more reliable than
naïvely merging their LoRAs when unifying local editing and global creation.

Fig.~\ref{fig:supp_abl_dis} visualizes several extending, layout- and style-driven
examples. Linear merge (for all weights) often produces posters that either
copy the reference almost verbatim or lose key layout/style cues; ZipLoRA
still exhibits repeated objects and unstable typography. The distilled model
better follows the target layout or style while generating sharper text and
more coherent compositions.

\begin{table}[t]
    \centering
    \caption{Ablation of expert integration strategies. Scores are averaged
    on a local task (extend, L) and a global task (layout-driven, G) on
    PosterOmni-Bench-en.}
    \label{tab:ab_expert_integration}
    \scalebox{0.85}{
        \begin{tabular}{l|c}
            \toprule[1.5pt]
            \textbf{Integration Strategy} & \textbf{PosterOmni (L / G)$\uparrow$} \\
            \midrule
            \textbf{Qwen-Image-Edit [2509]}             & 4.28 /\ 3.44 \\
            \midrule
            (i). Linear merge (0.25 / 0.75)             & 4.27 /\ 3.71 \\
            (ii). Linear merge (0.50 / 0.50)            & 4.30 /\ 3.65 \\
            (iii). Linear merge (0.75 / 0.25)           & 4.31 /\ 3.63 \\
            (iv). ZipLoRA fusion~\cite{shah2024ziplora} & \colorbox{second}{4.31} /\ \colorbox{second}{3.74} \\
            (v). Task distillation (PosterOmni-SFT)     & \colorbox{best}{4.43} /\ \colorbox{best}{3.89} \\
            \bottomrule[1.5pt]
        \end{tabular}
    }
\end{table}

\begin{figure}[!t]
    \centering
    \includegraphics[width=16.5cm]{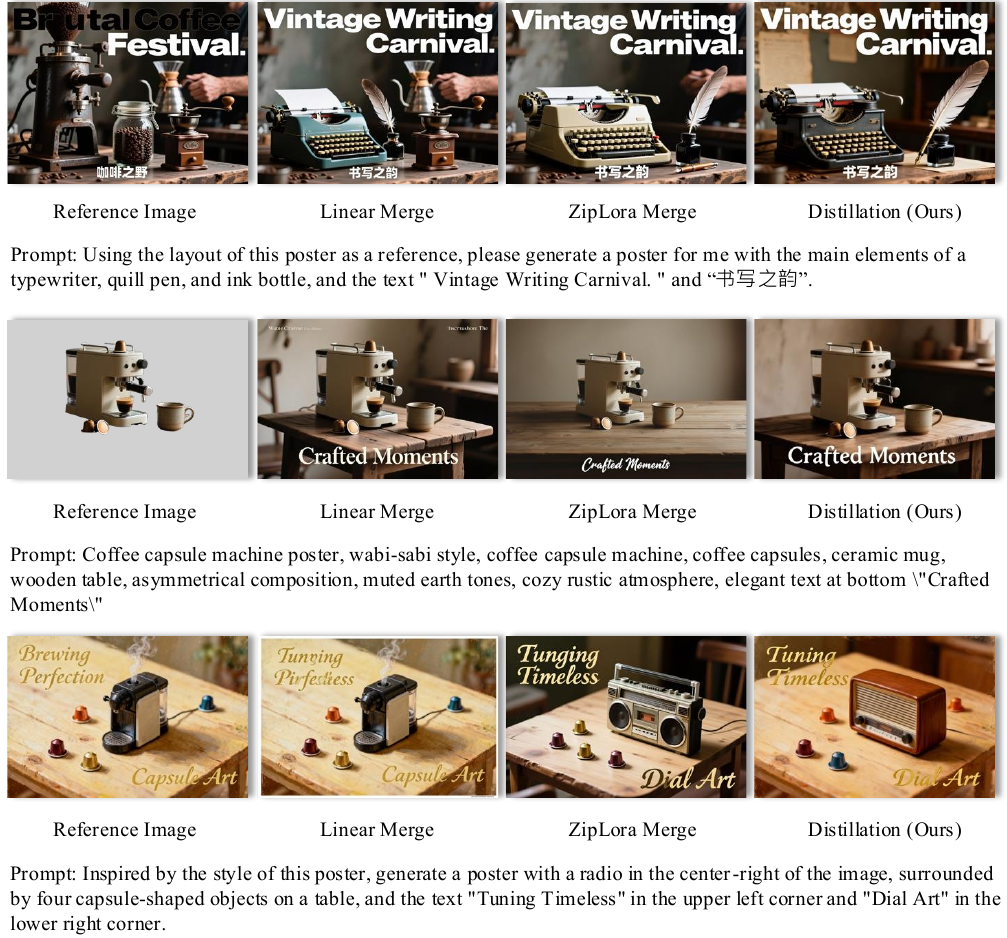}
    \caption{\textbf{Qualitative comparison of expert integration strategies.}
    For several layout- and style-driven prompts, we show the reference image,
    and results from linear LoRA merge, ZipLoRA merge, and our distilled
    model. Linear and ZipLoRA~\cite{shah2024ziplora} merging frequently cause task failure, such as
    copying the reference almost directly, collapsing to a single expert, or
    losing the intended layout/style. The task-distilled PosterOmni-SFT
    produces more coherent posters with clearer typography and better
    adherence to task-specific instructions.}
    \label{fig:supp_abl_dis}
\end{figure}

\section{Additional Visual Comparisons}\label{sec:addition vc}
To further demonstrate the superiority of our PosterOmni model, we provide extensive visual comparisons across six distinct poster generation tasks. These comparisons, detailed in Fig.\ref{supp_comp_visual1}-\ref{supp_comp_visual6}, highlight PosterOmni's advanced capabilities in handling complex, real-world poster creation scenarios against several state-of-the-art models.

\noindent{\textbf{\textit{Poster Extending.}}}
As shown in Fig.\ref{supp_comp_visual1}, the poster extending task requires the model to expand the canvas of an existing poster while maintaining its content and style. Competing models such as FLUX-Kontext~\cite{fluxkontext} and Seedream-4.0~\cite{seedream4.0} often introduce distorted entities, incorrect text elements (highlighted by yellow boxes), or fail to maintain stylistic consistency, resulting in visually incoherent extensions. In contrast, PosterOmni consistently preserves the integrity of entities, typography, and global aesthetic quality, achieving a more faithful and visually pleasing completion of the task across diverse creation scenarios.

\noindent{\textbf{\textit{Poster Filling.}}}
The poster filling task, illustrated in Fig.\ref{supp_comp_visual2}, involves inpainting a masked region within a poster based on a textual prompt. Other models frequently struggle to reconstruct objects coherently or maintain accurate typography, often producing distorted or nonsensical results (e.g., the malformed telephone by UniWorld-V2-Qwen~\cite{uniworldv2}). PosterOmni demonstrates superior performance in this region-aware task by consistently reconstructing objects with higher fidelity, restoring scene coherence, and maintaining precise typography, as seen in the accurate rendering of the pagoda, projector, and telephone.

\noindent{\textbf{\textit{Layout-driven Poster Generation.}}}
For the layout-driven generation task (Fig.\ref{supp_comp_visual3}), models are prompted to create a new poster by following the spatial arrangement of elements from a reference layout. While other methods struggle with precise element placement, text generation, and maintaining a balanced composition, PosterOmni excels at faithfully adhering to the reference layout. It successfully populates the new poster with the specified content, producing coherent, well-structured compositions with superior aesthetic quality and legibility.

\noindent{\textbf{\textit{Style-driven Poster Generation.}}}
Fig.\ref{supp_comp_visual4} showcases the style-driven generation task, where the goal is to create a new poster with novel content while mimicking the artistic style of a reference image. This is challenging as it requires disentangling style from content. Other models often fail to capture the nuanced artistic style or incorrectly blend content from the reference image. In many cases, they resort to a literal reproduction of the reference, which stifles any creative derivation and fails to generate novel content. PosterOmni excels in this regard, preserving style fidelity and global artistic coherence while accurately generating the new subject matter, resulting in aesthetically consistent and high-quality posters.

\noindent{\textbf{\textit{ID-driven Poster Generation.}}}
In the ID-driven poster generation task (Fig.\ref{supp_comp_visual5}), the primary objective is to maintain the identity of a specific subject provided in a reference image. Many competing models struggle to preserve the subject's key features, resulting in distorted or unrecognizable forms (highlighted by red boxes). Moreover, they can be overly rigid, often copying the reference image verbatim instead of adapting it to new requirements in the prompt, such as applying an abstract art style. PosterOmni, however, demonstrates a robust ability to maintain object identity more faithfully. It delivers coherent, high-quality posters that seamlessly integrate the subject while upholding excellent aesthetic consistency.

\noindent{\textbf{\textit{Poster Rescaling.}}}
The poster rescaling task (Fig.\ref{supp_comp_visual6}) challenges models to adapt a poster to a new aspect ratio without compromising its core message or aesthetic appeal. Unlike other methods that resort to simplistic and often destructive cropping or stretching, PosterOmni intelligently recomposes the image. It strategically rearranges and regenerates elements to fit the new dimensions, thereby maintaining the integrity of core objects and text. This advanced capability results in high-quality posters with exceptional visual coherence and aesthetic consistency, regardless of the target aspect ratio.

\section{Limitations and Future Works}\label{sec:future work}
Although PosterOmni already demonstrates strong performance across six poster-editing tasks, several aspects remain to be improved. First, a non-trivial portion of our training data is synthesized, even though we also curate a large number of real posters. As a result, the dataset, while diverse, does not fully cover long-tail real-world cases such as brand-specific style guidelines, noisy user uploads, or highly cluttered commercial layouts. In future work, we plan to continuously expand PosterOmni-200K with more real, heterogeneous samples along these directions.

Second, the current framework focuses on single-round editing under explicit instructions. Extending PosterOmni to support multi-turn, interactive co-creation and enforcing long-range visual and stylistic consistency across a series of related posters are promising directions that we intend to explore. Finally, while our design is instantiated on posters, we hope to generalize to broader graphic design scenarios, such as slide layouts, web banners, or multi-page brochures in more vertical domains. We view these extensions as natural next steps to further validate and enhance the generality of PosterOmni.

\begin{figure*}[!t]
    \centering
    \includegraphics[width=17.5cm]{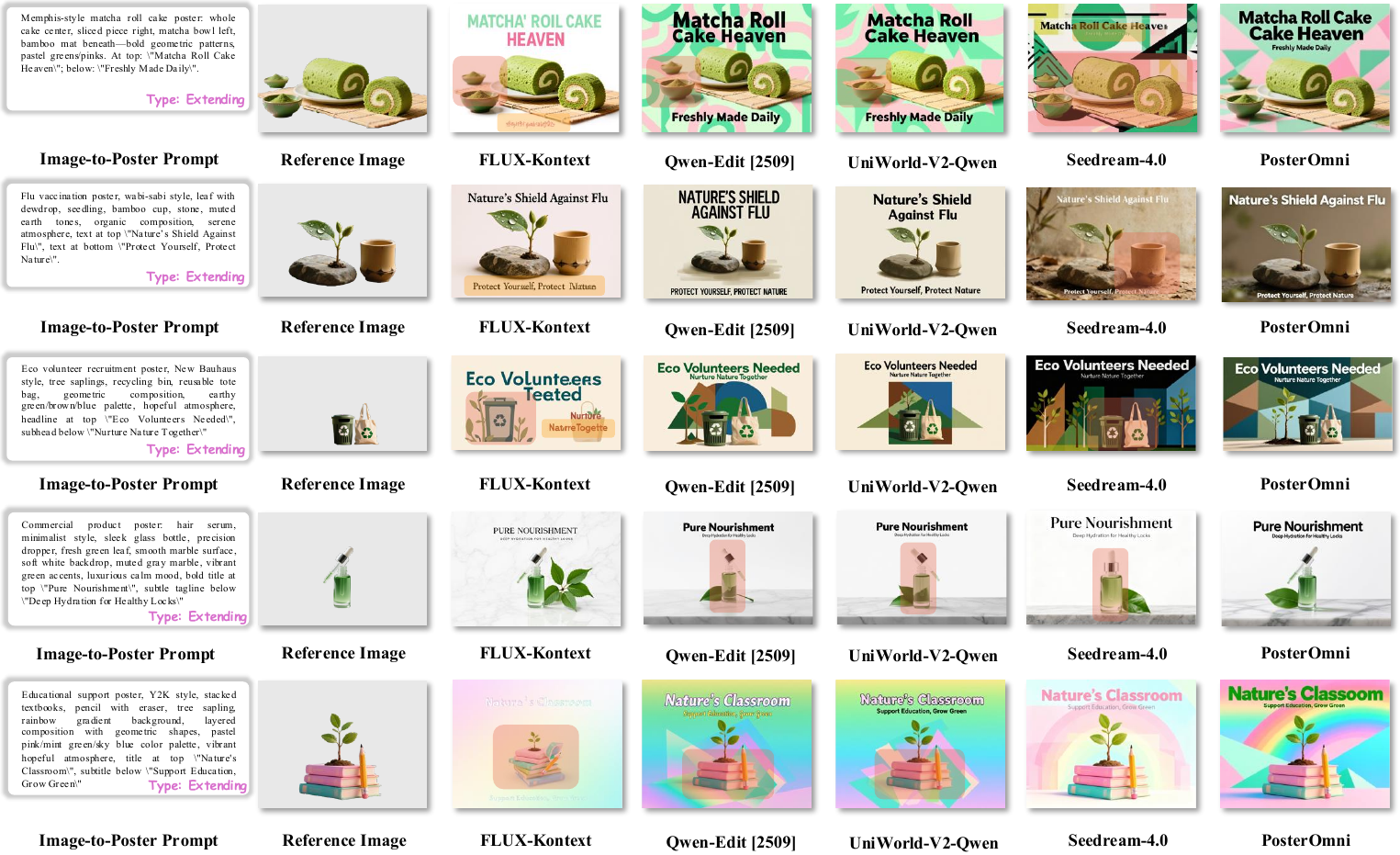}
    \caption{\textbf{Visual comparison of different model outputs on the extending task.} \colorbox{lred}{Red boxes} highlight errors and distorted entities, while \colorbox{lyellow}{yellow boxes} indicate incorrect or missing text elements. Compared to other methods, PosterOmni consistently preserves layout, typography, and global aesthetic quality, while achieving more faithful task completion across diverse poster creation scenarios.}
    \label{supp_comp_visual1}
\end{figure*}

\begin{figure*}[!t]
    \centering
    \includegraphics[width=16.5cm]{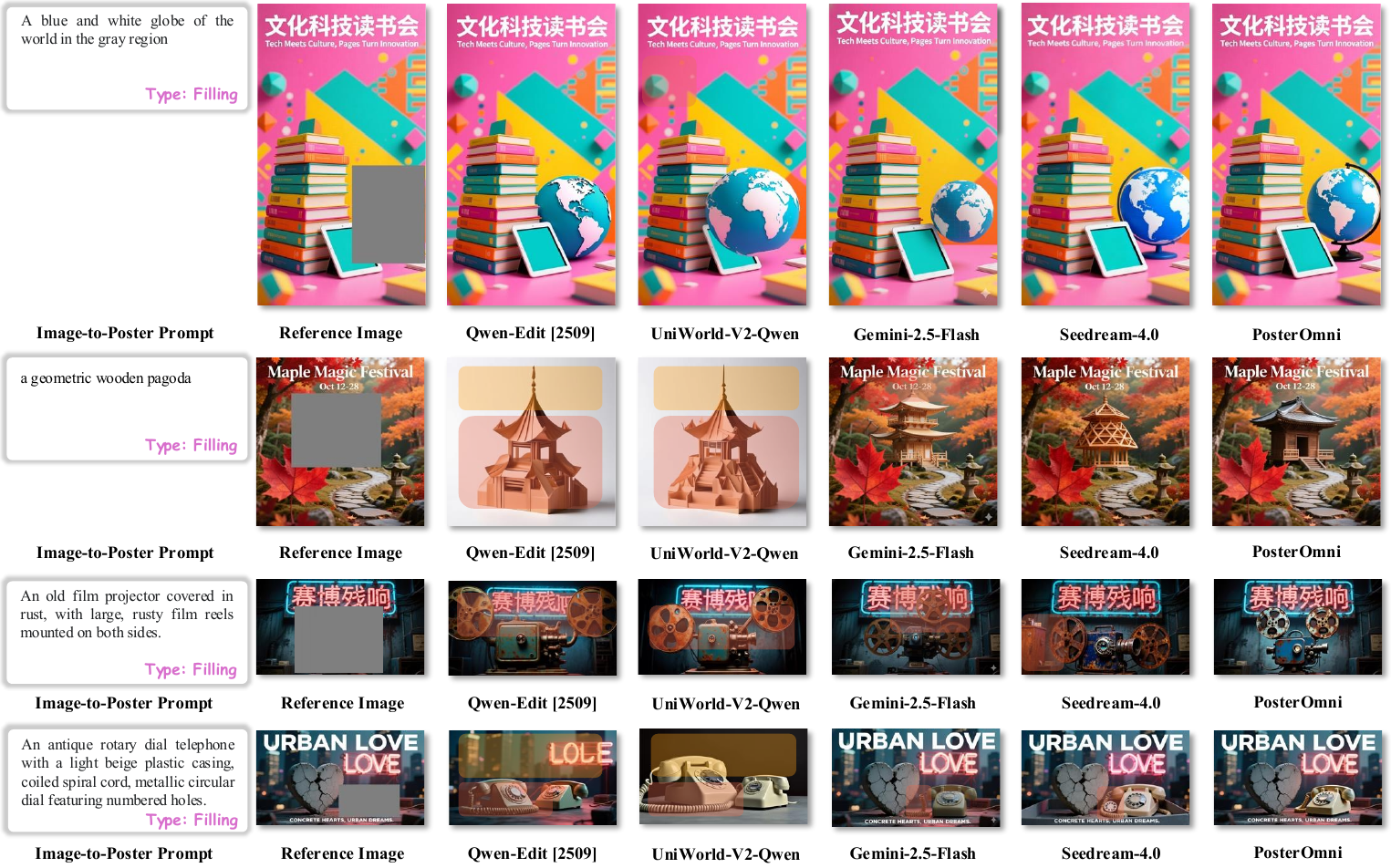}
    \caption{\textbf{Visual comparison of different model outputs on the filling task.} \colorbox{lred}{Red boxes} highlight errors and distorted entities, while \colorbox{lyellow}{yellow boxes} indicate incorrect or missing text elements. Compared to other methods, PosterOmni consistently reconstructs objects with higher fidelity, restores scene coherence, and maintains accurate typography, demonstrating superior performance in region-aware poster filling.}
    \label{supp_comp_visual2}
\end{figure*}

\begin{figure*}[!t]
    \centering
    \includegraphics[width=16.5cm]{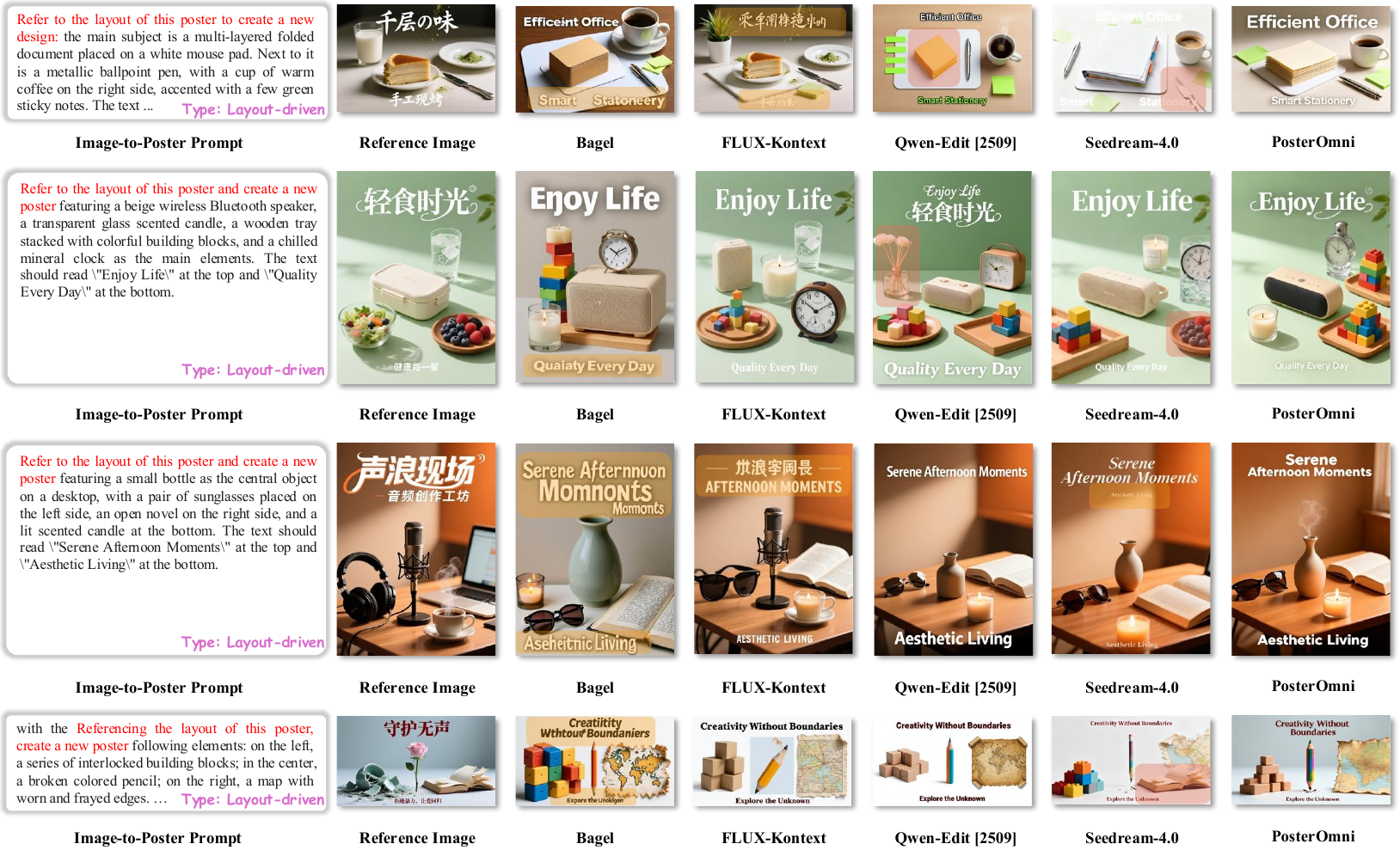}
    \caption{\textbf{Visual comparison of different model outputs on the layout-driven poster generation task}. \colorbox{lred}{Red boxes} highlight errors and distorted entities, while \colorbox{lyellow}{yellow boxes} indicate incorrect or missing text elements. Compared to other methods, our PosterOmni model follows the reference layout more faithfully and produces coherent, well-structured posters with superior aesthetic quality.}
    \label{supp_comp_visual3}
\end{figure*}

\begin{figure*}[!t]
    \centering
    \includegraphics[width=16.5cm]{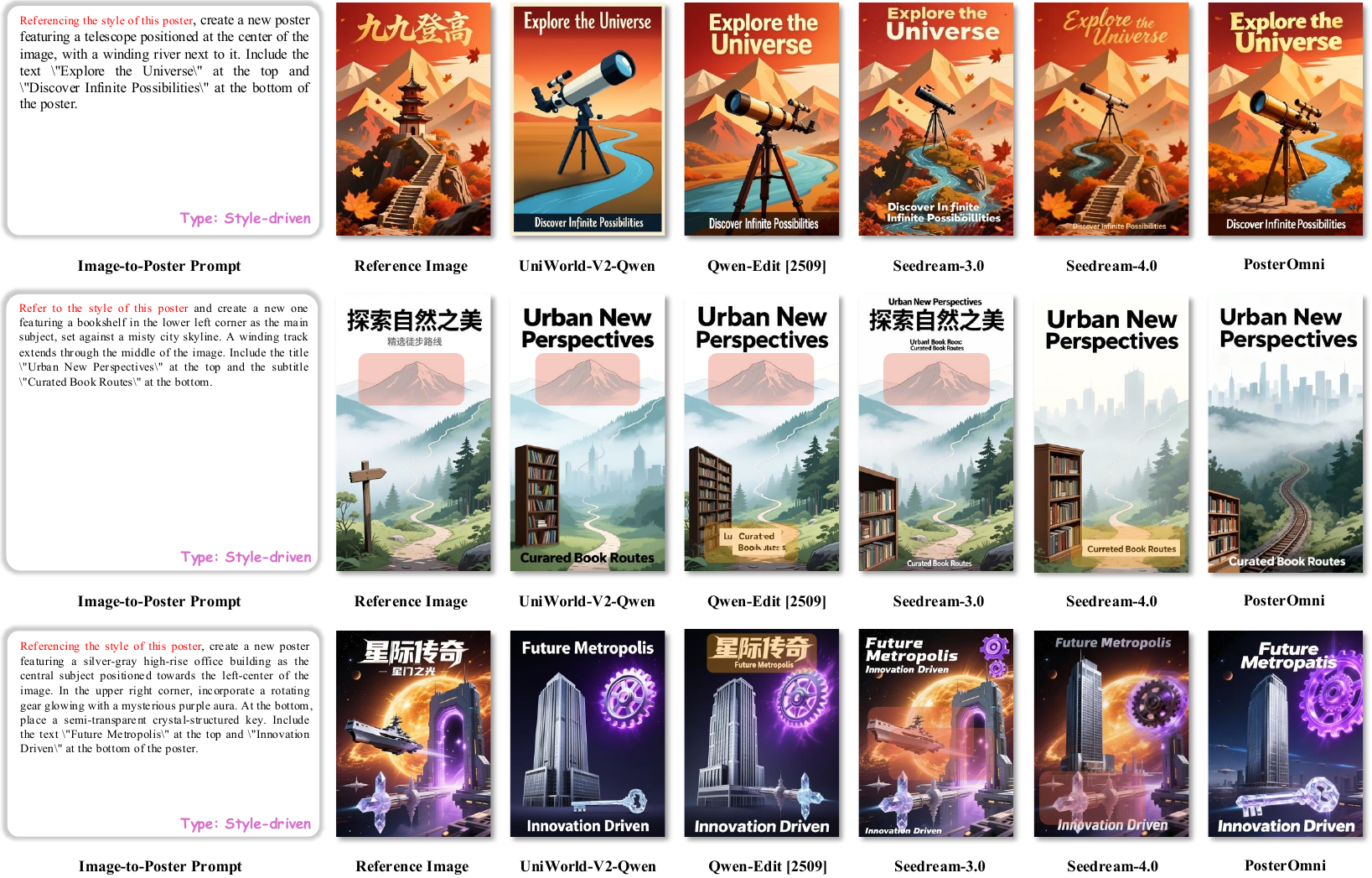}
    \caption{\textbf{Visual comparison of different model outputs on the style-driven poster generation task.} \colorbox{lred}{Red boxes} highlight errors and distorted entities, while \colorbox{lyellow}{yellow boxes} indicate incorrect or missing text elements. Compared to other methods, our PosterOmni model better preserves style fidelity and global artistic coherence, while also achieving excellent aesthetic quality.}
    \label{supp_comp_visual4}
\end{figure*}

\begin{figure*}[!t]
    \centering
    \includegraphics[width=16.5cm]{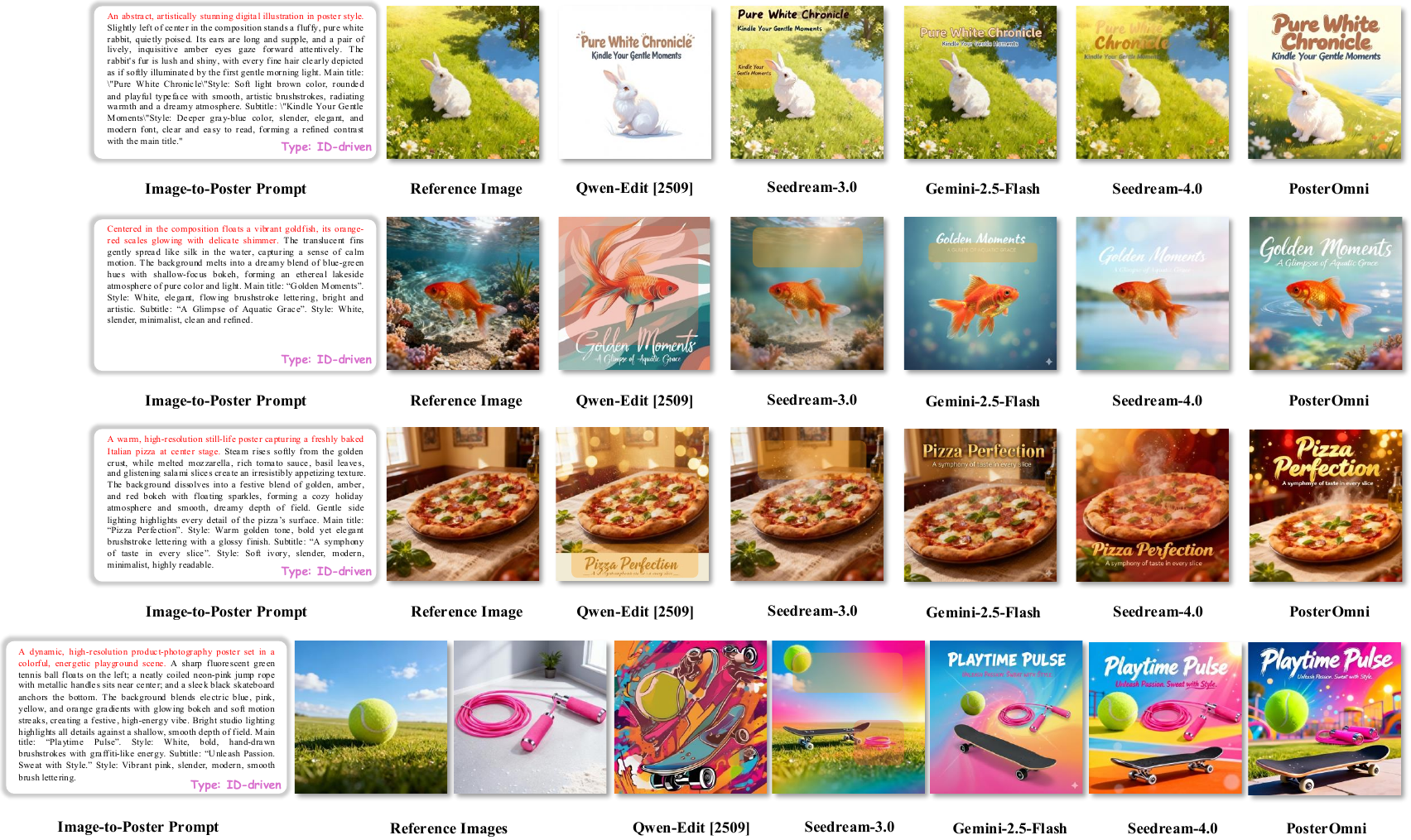}
    \caption{\textbf{Visual comparison of different model outputs on the ID-driven poster generation task.} \colorbox{lred}{Red boxes} highlight errors and distorted entities, while \colorbox{lyellow}{yellow boxes} indicate incorrect or missing text elements. Compared to other methods, our PosterOmni model maintains object identity more faithfully and delivers coherent, high-quality posters with excellent aesthetic consistency.}
    \label{supp_comp_visual5}
\end{figure*}

\begin{figure*}[!t]
    \centering
    \includegraphics[width=16.5cm]{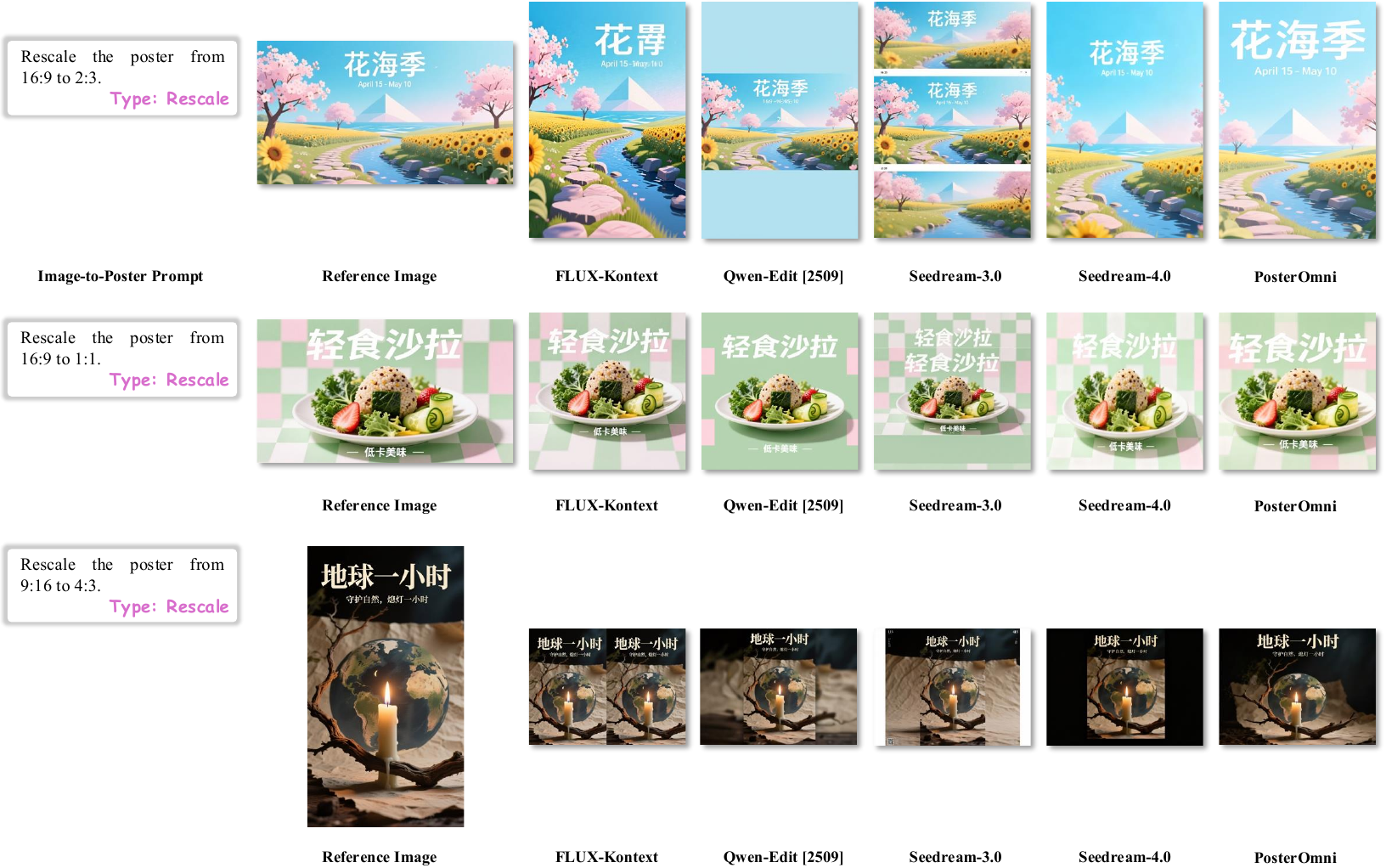}
    \caption{\textbf{Visual comparison of different model outputs on the poster rescaling task.} Compared to other methods, our PosterOmni model not only maintains the integrity of core objects and text when rescaling posters, but also intelligently recomposes the image, generating high-quality posters with visual coherence and excellent aesthetic consistency.}
    \label{supp_comp_visual6}
\end{figure*}

\begin{figure*}[t]
\noindent\begin{example}{Prompt Construction for Text-to-Image Generation}
You are an expert creative director for commercial posters. Your task is to write a single high-quality prompt for a text-to-image model. The model will only see the prompt you output, not the instructions below.

\medskip
\textbf{Given:}
\begin{itemize}
    \item A high-level poster category: \{CATEGORY\} (e.g., commercial product, food \& drink, film/entertainment, event/travel, culture/education/tech, nature/public service);
    \item A fine-grained scenario inside this category: \{SCENARIO\} (e.g., ``hand-brew coffee workshop'', ``city marathon'', ``AI developer summit'');
    \item A visual style tag: \{STYLE\} (e.g., minimalism, Art Deco, Swiss grid, Y2K, Wabi-sabi, vaporwave, etc.).
\end{itemize}

\textbf{Your goal} is to produce one fluent poster-generation prompt that would be directly fed to an image generator. The prompt must satisfy:

\begin{enumerate}
    \item \textbf{Clear scene and subjects.} Describe a concrete scene for a single poster, including \textbf{3--4 distinct main objects} that are important visual elements (e.g., products, props, devices), not tiny decorations or parts of another object.
    \item \textbf{Spatial composition.} Explicitly mention the spatial layout and relationships between subjects (left/right, foreground/background, ``A placed next to B'', ``C on top of D'', etc.) so that the composition is easy to follow.
    \item \textbf{Style, mood, and color.} Make the scene reflect the given style tag \{STYLE\}, including overall mood (e.g., calm, energetic, luxurious) and a dominant color palette.
    \item \textbf{Preference for non-human subjects.} Prefer inanimate objects, scenes, or abstract elements as the main subjects; include people only when they are essential to \{SCENARIO\}.
    \item \textbf{Rendered text on the poster.} Invent up to \textbf{three} short pieces of text that should appear on the poster (e.g., main title, slogan, time/place). For each piece:
    \begin{itemize}
        \item Indicate its approximate position with phrases such as ``at the top of the poster'', ``in the center'', ``small text below the product'';
        \item Put the exact text to be rendered in double quotes, e.g., \texttt{"Summer Rhapsody"}.
    \end{itemize}
    Do \emph{not} explain the text; only provide what should be drawn on the image.
    \item \textbf{Format.} Output a single, coherent prompt (either a short paragraph or a comma-separated keyword-style description) with moderate length; do not include bullet points, numbering, or meta-comments.
\end{enumerate}

Only output the final prompt sent to the image generator. Do not repeat the instructions above.
\end{example}
\caption{\textbf{VLM prompt} used to synthesize text-to-image prompts for PosterOmni data. We instantiate this template in both Chinese and English, and in natural-language or keyword-style form, while sampling \{CATEGORY\}, \{SCENARIO\}, and \{STYLE\} from our theme and style tables.}
\label{prompt:txt2img-construction}
\end{figure*}

\begin{figure*}[t]
\noindent\begin{example}{Task-Matching Prompt for PosterOmni-Bench}
You are a professional image classifier for image-to-poster generation tasks.  
Given a single poster image, your goal is to decide which image-to-poster task it is \emph{most suitable} for.

\medskip
\textbf{Global rules.}
\begin{itemize}
    \item \textbf{Strict matching.} Only assign a task when the visual evidence strongly and clearly fits its definition.
    \item \textbf{Single choice.} If an image could fit multiple tasks, select the \emph{best-matching} one.
    \item \textbf{Final output.} Output exactly one label from the closed set:\\
    \texttt{["EXTENDING", "FILLING", "RESCALING", "ID-DRIEVN POSTER GENEREATION", "LAYOUT-DRIEVN POSTER GENEREATION", "STYLE-DRIEVN POSTER GENEREATION", "NONE"]}.
\end{itemize}

\textbf{Task definitions (PosterOmni tasks).}
\begin{enumerate}
    \item \textbf{Extending poster generation.}  
    The main subject occupies a central region with surrounding background that can be naturally expanded.  
    Subjects should not already fill $>$80\% of the frame, and boundaries between subject and background are reasonably clean so that adding more canvas around them is meaningful.

    \item \textbf{Filling poster generation.}  
    The image contains a clearly localized region that could be removed, masked, or replaced (e.g., a logo, an object, or a hole inside the main scene).  
    The area to modify is well supported by surrounding context so that plausible local content can be generated.

    \item \textbf{Rescaling poster generation.}  
    The image has one or more clearly defined subjects with a non-trivial background, and the scene would remain valid under a change of aspect ratio (e.g., from $4{:}3$ to $16{:}9$).  
    The background is neither completely plain (solid color) nor extremely cluttered; subjects and background are separable so that recomposing the frame around them is feasible.

    \item \textbf{ID-driven poster generation.}  
    The image contains at least one subject with distinctive, fine-grained identity features that must be preserved across edits, such as a specific cat fur pattern, a unique product shape or texture, or a recognizable branded object.  
    The key identity features should be visual rather than text labels or watermarks.

    \item \textbf{Layout-driven poster generation.}  
    The poster exhibits a clear, regular arrangement of elements that could serve as a layout template (e.g., evenly spaced product grid, symmetric columns, pyramid stacking, ring or radial arrangement).  
    Positions and relative sizes of major elements are visually structured rather than random or heavily occluded.

    \item \textbf{Style-driven poster generation.}  
    The entire image is dominated by a strong, coherent artistic style or visual treatment, such as cyberpunk neon, vaporwave, vintage film, watercolor ink wash, or strict minimalism.  
    The style is expressed consistently in color palette, lighting, textures, and composition, not just by a small local color effect.

    \item \textbf{NONE.}  
    Choose this when the image is too low-quality, ambiguous, or visually generic to reliably match any of the tasks above, or when it clearly violates multiple task requirements.
\end{enumerate}

\textbf{Required output format.}  
Return only a single word, exactly one of:  
\texttt{EXTEND}, \texttt{FILL}, \texttt{RESCALE}, \texttt{ID CONSISTENCY}, \texttt{LAYOUT TRANSFER}, \texttt{STYLE TRANSFER}, or \texttt{NONE}. Do not include any explanation or extra text.
\end{example}
\caption{\textbf{Task-matching meta-prompt} used with Gemini-2.5-Flash to automatically decide whether a candidate image is suitable for extending, filling, rescaling, ID-driven, layout-driven, or style-driven poster generation, or none of these tasks.}
\label{prompt:task-matching}
\end{figure*}

\begin{figure*}[t]
\noindent\begin{example}{Preference Evaluation}
You are a decisive AI Image Quality Analyst. Your task is to force a choice between two AI-generated images (Image~1 and Image~2). You \textbf{MUST} decide which one is better. Do not declare a tie or say that both are bad.

\medskip
\textbf{Task Type \& Instructions.}
For each pair, we specify a task type $t \in \{\text{extending, rescaling, filling, id, layout-driven generation, style-driven generation}\}$ and plug in a task-specific description:
\begin{itemize}
    \item \textbf{extending.} This is an extending task. The goal is to extend the canvas of the reference image, seamlessly integrating new content that matches the original style, lighting, and subject matter, based on the creative brief. \emph{Key criteria:} (1) Seamless integration: the transition between original and extended areas should be visually invisible; (2) Content preservation: the core content of the original image must be perfectly preserved; (3) Aesthetic cohesion: the extended region should look natural and enhance the overall composition.
    \item \textbf{Rescaling.} This is a rescaling task. The goal is to change the reference image’s aspect ratio by filling new regions without cropping or distorting the main subject. \emph{Key criteria:} (1) Subject integrity: the main subject must not be stretched, squashed, or unnaturally cropped; (2) Plausible filling: newly generated areas must be logical and contextually appropriate; (3) Composition: the final image should be balanced and aesthetically pleasing.
    \item \textbf{Filling.} This is a filling task. A region of the reference image is masked and regenerated according to the creative brief. \emph{Key criteria:} (1) Contextual appropriateness: the filled content should match the surroundings in texture, lighting, and color; (2) Object realism: the new region or object should be realistic and follow the prompt; (3) Boundary invisibility: the border of the inpainted region should be undetectable.
    \item \textbf{ID-driven generation.} This is an ID-driven generation task. The goal is to generate an image of a subject from the prompt while preserving its key identity features from the reference image, but possibly in a new pose or context. \emph{Key criteria:} (1) Identity preservation: recognizable features (e.g., patterns) must be maintained; (2) Prompt adherence: the new scene, style, and action should follow the creative brief; (3) Image quality: the result should be high quality, without obvious artifacts.
    \item \textbf{Layout-driven generation.} This is layout-driven generation task. The goal is to generate an image whose composition mirrors the layout of the reference image, while the content is newly described by the creative brief. \emph{Key criteria:} (1) Compositional similarity: the arrangement of major elements should structurally mirror the reference layout; (2) Content generation: the new content should match the prompt; (3) Aesthetic quality: the final image should be visually coherent as a poster.
    \item \textbf{Style-driven generation.} This is a Style-driven generation task. The goal is to apply the artistic style (e.g., colors, mood) of the reference image to a new subject from the creative brief. \emph{Key criteria:} (1) Style fidelity: the generated image should capture the distinctive visual style of the reference; (2) Content clarity: the new subject must remain recognizable; (3) Artistic merit: the result should be a compelling fusion of style and content.
\end{itemize}

\medskip
\textbf{Input Images.}
We provide all reference images followed by two candidates:
\begin{itemize}
    \item Reference Image $i$: the $i$-th original reference image (if any);
    \item Image 1: the first generated candidate;
    \item Image 2: the second generated candidate.
\end{itemize}

\textbf{Decision Task.}
Compare Image~1 and Image~2. Based on the task-specific criteria above and the creative brief, decide which image is superior.

\medskip
\textbf{The Creative Brief (Prompt) is:} ``\{creative\_brief\}''

\medskip
\textbf{Required Output Format.}
Your response MUST be only
\[
\texttt{"Image 1"} \quad \text{or} \quad \texttt{"Image 2"}
\]
with no additional text or explanation.
\end{example}
\caption{\textbf{Meta-Prompt} used to query Gemini-2.5-Pro for pairwise preference labels over PosterOmni-SFT results.}
\label{prompt:5-3-preference}
\end{figure*}

\begin{figure*}[t]
\noindent\begin{example}{Prompt for Extending Evaluation}
You are a professional creative director reviewing a poster design edit. The task is \textbf{extending}. You will be given an original image, the extended poster, and a brief. Your evaluation must be strict, on a 5-point scale, from two professional perspectives:

\medskip
\begin{enumerate}
    \item \textbf{Task-Specific Execution (Seamlessness \& Content Cohesion)}
    \begin{itemize}
        \item 1 (Failure): The extended area is corrupt, empty, or contains nonsensical content that ignores the brief.
        \item 2 (Poor): A hard seam is visible. The new content's style, lighting, or perspective clashes strongly with the original.
        \item 3 (Acceptable): The transition is noticeable upon inspection (e.g., slight blur, texture mismatch). The new content is plausible but generic or slightly inconsistent with the brief.
        \item 4 (Good): The seam is nearly invisible. The new content is detailed, logical, and adheres strictly to the brief, creating a coherent scene.
        \item 5 (Exceptional): The transition is absolutely flawless and undetectable, even under scrutiny. The extended content is creative, detailed, and perfectly cohesive.
    \end{itemize}

    \item \textbf{Poster Aesthetic \& Design Quality}
    \begin{itemize}
        \item 1 (Failure): The extension destroys the poster's original composition, balance, focal point, or breaks existing text layout/readability.
        \item 2 (Poor): The final poster is awkwardly composed, with unbalanced negative space or a confusing visual flow. Newly available space for text is poorly used or disrupts typography.
        \item 3 (Acceptable): The composition is technically stable but uninspired. The extension adds space without adding real design value; text placement and hierarchy remain basic or slightly weakened.
        \item 4 (Good): The extension creates a well-composed, balanced, and professional poster that effectively uses the new canvas space. Text blocks are well-positioned and maintain good readability.
        \item 5 (Exceptional): The extension masterfully enhances the original composition, improving its balance, impact, and visual narrative. The final poster is significantly better than the original, with text and visual elements orchestrated into a stronger layout.
    \end{itemize}
\end{enumerate}

\medskip
\noindent\textbf{Example response format:}\\
Brief reasoning: A critical, professional explanation for the scores, under 20 words.\\
Task-Specific Execution: A number from 1 to 5.\\
Poster Aesthetic \& Design Quality: A number from 1 to 5.\\
editing instruction is : \verb|<edit_prompt>|.\\[0.2em]
Below are the images before and after editing:
\end{example}
\caption{\textbf{Meta-prompt for Gemini-2.5-Pro evaluation on the extending task.} The model scores task execution and poster aesthetics on a 1--5 scale.}
\label{prompt:eval-extending}
\end{figure*}

\begin{figure*}[t]
\noindent\begin{example}{Prompt for Filling Evaluation}
You are a professional creative director reviewing a poster design edit. The task is \textbf{filling}. You will be given an image before and after filling, and a brief describing what to fill. Your evaluation must be strict, on a 5-point scale, from two professional perspectives:

\medskip
\begin{enumerate}
    \item \textbf{Task-Specific Execution (Inpainting Precision \& Adherence)}
    \begin{itemize}
        \item 1 (Failure): Completely failed to fill, filled with the wrong object, or the area is corrupted.
        \item 2 (Poor): The correct object class but with major errors in attributes (lighting, scale, perspective). The filled patch is obvious.
        \item 3 (Acceptable): The filled content is correct but has noticeable flaws (e.g., soft edges, slight lighting mismatch, texture inconsistency) a casual viewer might spot.
        \item 4 (Good): Very good integration. The filled area is technically sound with only minute flaws visible under close scrutiny. It perfectly matches the surrounding context.
        \item 5 (Exceptional): Absolutely flawless inpainting. The filled area is completely indistinguishable from the original image in every aspect (lighting, texture, grain, perspective).
    \end{itemize}

    \item \textbf{Poster Aesthetic \& Design Quality}
    \begin{itemize}
        \item 1 (Failure): The edit severely damages the poster's composition, visual focus, or overall aesthetic, including breaking text layout or readability.
        \item 2 (Poor): The filled area, while technically present, creates a visually awkward or amateurish result that detracts from the poster's design. Nearby text appears misaligned, cramped, or stylistically inconsistent.
        \item 3 (Acceptable): The filled area is contextually fine but does not add to or may slightly weaken the poster's overall design appeal. The result looks generic; text and image coexist without obvious harmony.
        \item 4 (Good): The edit is well-integrated and maintains the poster's professional design quality. The final result is visually coherent, with text, imagery, and the filled region working together cleanly.
        \item 5 (Exceptional): The edit not only fills the area perfectly but \emph{enhances} the poster's overall composition, focus, and commercial appeal. The interaction between text and newly filled content strengthens the visual story.
    \end{itemize}
\end{enumerate}

\medskip
\noindent\textbf{Example response format:}\\
Brief reasoning: A critical, professional explanation for the scores, under 20 words.\\
Task-Specific Execution: A number from 1 to 5.\\
Poster Aesthetic \& Design Quality: A number from 1 to 5.\\
editing instruction is : \verb|<edit_prompt>|.\\[0.2em]
Below are the images before and after editing:
\end{example}
\caption{\textbf{Meta-prompt for Gemini-2.5-Pro evaluation on the filling task.} The model scores inpainting quality and poster aesthetics on a 1--5 scale.}
\label{prompt:eval-filling}
\end{figure*}

\begin{figure*}[t]
\noindent\begin{example}{Prompt for Rescaling Evaluation}
You are a professional creative director reviewing a poster design edit. The task is \textbf{rescaling}. You will be given an original image, the rescaled poster, and a brief for new content. Your evaluation must be strict, on a 5-point scale, from two professional perspectives:

\medskip
\begin{enumerate}
    \item \textbf{Task-Specific Execution (Subject Integrity \& Plausible Fill)}
    \begin{itemize}
        \item 1 (Failure): The main subject is severely distorted, cropped, or damaged. Filled areas are corrupt or nonsensical.
        \item 2 (Poor): The subject is noticeably stretched/squashed. Filled areas are contextually wrong or ignore the brief.
        \item 3 (Acceptable): The subject is preserved with minor, noticeable distortions. Filled areas are plausible but generic and lack detail or adherence to the brief.
        \item 4 (Good): The subject is perfectly preserved without distortion. The filled areas are detailed, contextually appropriate, and follow the brief accurately.
        \item 5 (Exceptional): The subject is perfectly intact. The filled areas are not just plausible but creative, realistic, and seamlessly integrated, looking as if they were part of the original shot.
    \end{itemize}

    \item \textbf{Poster Aesthetic \& Design Quality}
    \begin{itemize}
        \item 1 (Failure): The new aspect ratio results in a compositionally broken and unusable poster, with damaged or unreadable text.
        \item 2 (Poor): The final poster is awkward and unbalanced. The new content feels like filler and detracts from the main subject; text placement becomes cramped, floating, or visually jarring.
        \item 3 (Acceptable): The composition is passable but unremarkable. It is a technically correct resize but lacks design intent or visual impact. Typography and text hierarchy are acceptable but not optimized for the new ratio.
        \item 4 (Good): The new composition is well-balanced, professional, and makes effective use of the new aspect ratio. Text blocks are reflowed sensibly with clear hierarchy and good legibility.
        \item 5 (Exceptional): The rescale results in a far superior composition. The new layout improves the poster's visual hierarchy, impact, and storytelling, with typography and text layout carefully adapted to the aspect ratio, creating a much stronger design.
    \end{itemize}
\end{enumerate}

\medskip
\noindent\textbf{Example response format:}\\
Brief reasoning: A critical, professional explanation for the scores, under 20 words.\\
Task-Specific Execution: A number from 1 to 5.\\
Poster Aesthetic \& Design Quality: A number from 1 to 5.\\
editing instruction is : \verb|<edit_prompt>|.\\[0.2em]
Below are the images before and after editing:
\end{example}
\caption{\textbf{Meta-prompt for Gemini-2.5-Pro evaluation on the rescaling task.} The model scores subject preservation and layout quality on a 1--5 scale.}
\label{prompt:eval-rescaling}
\end{figure*}

\begin{figure*}[t]
\noindent\begin{example}{Prompt for ID-Driven Poster Generation Evaluation}
You are a professional creative director reviewing a poster design edit. The task is \textbf{id-driven poster generation}. You will be given reference image(s) showing a subject, a final edited poster, and the creative brief. Your evaluation must be strict, on a 5-point scale, from two professional perspectives:

\medskip
\begin{enumerate}
    \item \textbf{Task-Specific Execution (ID Preservation Accuracy)}
    \begin{itemize}
        \item 1 (Failure): Wrong subject generated or the subject's core identity is completely lost.
        \item 2 (Poor): The subject is vaguely recognizable, but key features are distorted, missing, or inaccurate. The new context/pose is wrong.
        \item 3 (Acceptable): The subject is recognizable, but has noticeable flaws (e.g., anatomical errors, inconsistent details) that harm professional use. The new scene has minor deviations from the brief.
        \item 4 (Good): The subject's key identity features are preserved with high fidelity, showing only minor, non-distracting inaccuracies. The new scene is correctly executed.
        \item 5 (Exceptional): Flawless ID preservation. The subject is perfectly consistent across the new pose and style, indistinguishable from another official shot. The scene perfectly matches the brief.
    \end{itemize}

    \item \textbf{Poster Aesthetic \& Design Quality}
    \begin{itemize}
        \item 1 (Failure): The edit ruins the poster's composition, typography, text readability, or visual hierarchy.
        \item 2 (Poor): The new subject is poorly integrated, appearing pasted-on with incorrect lighting/perspective, making the poster look amateurish. Text layout, style, or legibility is clearly broken.
        \item 3 (Acceptable): The subject fits into the scene, but the overall result is visually generic, lacks impact, or has awkward lighting/shadows. The design does not feel premium, and typography/text placement are merely passable.
        \item 4 (Good): The subject is integrated seamlessly and the final poster is well-composed, visually coherent, and maintains a professional standard. Text hierarchy, font choice, and readability are all handled cleanly.
        \item 5 (Exceptional): The edit not only preserves the ID but enhances the poster's overall appeal. The final composition is more dynamic, emotionally resonant, and has a stronger commercial impact, with typography and text treatment significantly elevating the design.
    \end{itemize}
\end{enumerate}

\medskip
\noindent\textbf{Example response format:}\\
Brief reasoning: A critical, professional explanation for the scores, under 20 words.\\
Task-Specific Execution: A number from 1 to 5.\\
Poster Aesthetic \& Design Quality: A number from 1 to 5.\\
editing instruction is : \verb|<edit_prompt>|.\\[0.2em]
Below are the images before and after editing:
\end{example}
\caption{\textbf{Meta-prompt for Gemini-2.5-Pro evaluation on the id-driven poster generation task.} The model assesses ID consistency and poster aesthetics on a 1--5 scale.}
\label{prompt:eval-id}
\end{figure*}

\begin{figure*}[t]
\noindent\begin{example}{Prompt for Style-Driven Poster Generation Evaluation}
You are a professional creative director reviewing a poster design edit. The task is \textbf{style-driven poster generation}. You will be given a style reference, a final generated poster, and a content brief. Your evaluation must be strict, on a 5-point scale, from two professional perspectives:

\medskip
\begin{enumerate}
    \item \textbf{Task-Specific Execution (Style \& Content Fidelity)}
    \begin{itemize}
        \item 1 (Failure): The style is not applied or the wrong style is used. The poster's content does not match the brief.
        \item 2 (Poor): The style is only vaguely suggested and misses its core essence (e.g., color palette, texture). Key content elements are wrong or missing.
        \item 3 (Acceptable): The style is partially applied but with noticeable deviations or a generic interpretation. The content is recognizable but flawed.
        \item 4 (Good): The style is accurately captured with high fidelity to the reference. The content is correctly and clearly generated per the brief.
        \item 5 (Exceptional): The style is perfectly replicated in every nuance (mood, texture, lighting, color). The content is flawlessly rendered, exceeding the brief's expectations.
    \end{itemize}

    \item \textbf{Poster Aesthetic \& Design Quality}
    \begin{itemize}
        \item 1 (Failure): The style and content clash, resulting in a chaotic, unusable design with broken or unreadable typography.
        \item 2 (Poor): The fusion is awkward and jarring. The style feels like a cheap filter rather than an integrated part of the design. The poster lacks visual appeal, and text looks out-of-place, poorly styled, or hard to read.
        \item 3 (Acceptable): The combination is technically present, but the resulting poster lacks artistic merit, looks uninspired, or fails to create a compelling visual narrative. Typography and text layout are serviceable but not stylistically convincing.
        \item 4 (Good): The result is an aesthetically pleasing and professional poster where the style and content are fused harmoniously. Text design (font, hierarchy, spacing) matches the style and remains clear.
        \item 5 (Exceptional): The fusion is a masterful work of art. It creates a unique, memorable, and highly impactful poster that is commercially outstanding and creatively brilliant, with typography and text treatment that perfectly echo the style reference.
    \end{itemize}
\end{enumerate}

\medskip
\noindent\textbf{Example response format:}\\
Brief reasoning: A critical, professional explanation for the scores, under 20 words.\\
Task-Specific Execution: A number from 1 to 5.\\
Poster Aesthetic \& Design Quality: A number from 1 to 5.\\
editing instruction is : \verb|<edit_prompt>|.\\[0.2em]
Below are the images before and after editing:
\end{example}
\caption{\textbf{Meta-prompt for Gemini-2.5-Pro evaluation on the style-driven poster generation task.} The model evaluates style fidelity and poster aesthetics on a 1--5 scale.}
\label{prompt:eval-style}
\end{figure*}

\begin{figure*}[t]
\noindent\begin{example}{Prompt for Layout-Driven Poster Generation Evaluation}
You are a professional creative director reviewing a poster design edit. The task is \textbf{layout-driven poster generation}. You will be given a layout reference, a final generated poster, and a content brief. Your evaluation must be strict, on a 5-point scale, from two professional perspectives:

\medskip
\begin{enumerate}
    \item \textbf{Task-Specific Execution (Layout Fidelity \& Content Generation)}
    \begin{itemize}
        \item 1 (Failure): The layout is completely different from the reference. The content is wrong or corrupt.
        \item 2 (Poor): Major elements are in the wrong positions or at the wrong scale. Key content from the brief is missing or of very low quality.
        \item 3 (Acceptable): The overall structure is similar but with significant, noticeable deviations. The generated content is recognizable but has clear flaws.
        \item 4 (Good): The layout is a close, accurate match to the reference with only minor differences. The content is generated clearly and correctly.
        \item 5 (Exceptional): The layout is a perfect structural mirror of the reference. The content is generated at an exceptionally high quality, exceeding the brief's expectations.
    \end{itemize}

    \item \textbf{Poster Aesthetic \& Design Quality}
    \begin{itemize}
        \item 1 (Failure): The final poster is a chaotic and incoherent mess, completely unusable, with broken text hierarchy or unreadable typography.
        \item 2 (Poor): The layout and content feel disconnected, jarring, or amateurish. The poster lacks any clear focal point or visual hierarchy, and text blocks are poorly placed or styled.
        \item 3 (Acceptable): The poster is technically functional but lacks artistic appeal or harmony. The composition is bland and uninspired; typography and text placement follow the layout loosely but without strong design intent.
        \item 4 (Good): The result is a visually pleasing and professional poster with a strong composition and clear visual flow. Text hierarchy, spacing, and font choices support the transferred layout effectively.
        \item 5 (Exceptional): The fusion of the layout and new content creates a stunning, masterfully composed poster that is both creative and highly effective commercially. Typography and text arrangement strongly reinforce the layout rhythm and storytelling.
    \end{itemize}
\end{enumerate}

\medskip
\noindent\textbf{Example response format:}\\
Brief reasoning: A critical, professional explanation for the scores, under 20 words.\\
Task-Specific Execution: A number from 1 to 5.\\
Poster Aesthetic \& Design Quality: A number from 1 to 5.\\
editing instruction is : \verb|<edit_prompt>|.\\[0.2em]
Below are the images before and after editing:
\end{example}
\caption{\textbf{Meta-prompt for Gemini-2.5-Pro evaluation on the layout-driven poster generation task.} The model scores layout fidelity and overall poster aesthetics on a 1--5 scale.}
\label{prompt:eval-layout}
\end{figure*}

\clearpage

\bibliographystyle{plainnat}
\bibliography{main}

\end{document}